\PassOptionsToPackage{table}{xcolor}
\documentclass[11pt]{article}
\usepackage[letterpaper,margin=1in]{geometry}
\usepackage{arena-brand}
\usepackage{natbib}
\setcitestyle{authoryear,round,citesep={;},aysep={,},yysep={;}}

\usepackage{hyperref}       
\usepackage{xurl}           
\usepackage{booktabs}       
\usepackage{amsmath}        
\usepackage{amsfonts}       
\usepackage{nicefrac}       
\usepackage{microtype}      
\usepackage{xcolor}         
\usepackage{graphicx}       
\usepackage{subcaption}     
\usepackage{framed}         
\usepackage{fancyvrb}       
\usepackage{wrapfig}        
\usepackage{caption}        
\usepackage{titlesec}
\usepackage{paralist}       
\usepackage[breakable,skins]{tcolorbox}

\colorlet{shadecolor}{black!5}
\fvset{fontsize=\scriptsize, frame=single, framesep=3pt, xleftmargin=2pt, xrightmargin=2pt}

\titleformat{\section}{\ArenaHeadingFont\color{ArenaInk}\fontsize{16}{19}\selectfont}{\thesection.}{0.6em}{}
\titleformat{\subsection}{\ArenaHeadingFont\color{ArenaInk}\fontsize{13.5}{16}\selectfont}{\thesubsection.}{0.6em}{}
\titleformat{\subsubsection}{\ArenaHeadingFont\color{ArenaInk}\fontsize{12}{14.5}\selectfont\bfseries}{\thesubsubsection.}{0.6em}{}
\titlespacing*{\section}{0pt}{2.6ex plus 0.5ex minus 0.2ex}{1.0ex}
\titlespacing*{\subsection}{0pt}{2.0ex plus 0.4ex minus 0.2ex}{0.7ex}
\titlespacing*{\paragraph}{0pt}{1.25ex}{1em}
\title{Arena-T2I Hard: Benchmarking and Improving Faithfulness with Dependency-Aware Checklist}

\author{%
  \mbox{Yuanhao Ban\textsuperscript{1,2}},
  \mbox{Tong Xie\textsuperscript{2}},
  \mbox{Sohyun An\textsuperscript{2}},
  \mbox{Yunqi Hong\textsuperscript{2}},
  \mbox{Evan Frick\textsuperscript{1}},
  \mbox{I-Hung Hsu\textsuperscript{1}},
  \mbox{Wei-Lin Chiang\textsuperscript{1}},
  \mbox{Ion Stoica\textsuperscript{1}},
  \mbox{Cho-Jui Hsieh\textsuperscript{1,2}}
}

\newcommand{\affiliations}{%
  \mbox{\textsuperscript{1}\,\href{https://arena.ai}{Arena Intelligence Inc}}\quad
  \mbox{\textsuperscript{2}\,\href{https://ucla.edu}{UCLA}}\\[4pt]
  \texttt{yuanhao@arena.ai, chojui@arena.ai}
}

\makeatletter
\renewcommand{\@maketitle}{%
  {\raggedright\setlength{\parindent}{0pt}%
    \vspace*{-0.5in}%
    \hbox to \linewidth{\ArenaLogo[0.256in]\hfil
      \raisebox{3pt}{\ttfamily\fontsize{9}{11}\selectfont\color{ArenaTextSecondary}\today}}\par
    \vspace{8pt}
    {\color{ArenaInk}\rule{\linewidth}{1pt}}\par
    \vspace{16pt}
    {\ArenaTitleFont\fontsize{28}{33}\selectfont\color{ArenaInk}\@title\par}
    \vspace{12pt}
    {\fontsize{11.5}{14}\selectfont\@author\par}
    \vspace{6pt}
    {\fontsize{9.5}{11.5}\selectfont\hypersetup{urlcolor=ArenaInk}\affiliations\par}%
  }
}
\makeatother
\renewenvironment{abstract}{%
  {\noindent\ArenaHeadingFont\large\bfseries Abstract\par}
  \vspace{8pt}\normalfont\normalsize
  \setlength{\parindent}{0pt}%
}{\par}

\begin{document}

\suppressfloats[t] 

\maketitle

\begin{tcolorbox}[enhanced, colback=ArenaPaper, colframe=ArenaBorderFaint,
  boxrule=0.75pt, arc=6pt, boxsep=0pt,
  left=12pt, right=12pt, top=12pt, bottom=12pt,
  before skip=16pt, after skip=12pt]
\begin{abstract}
Faithfulness---how precisely a generated image aligns with its
prompt---is increasingly central to the real-world utility of
text-to-image (T2I) models. Existing faithfulness benchmarks, however,
rely on simple atomic instructions, on which top-tier systems already
achieve near-perfect scores. As T2I models enter creative workflows, users issue multi-faceted requests combining intricate spatial relationships,
stylistic constraints, and complex text rendering. In this setting, a single
binary VLM-judge score no longer captures \emph{which} specific
constraints the model fails to satisfy.
We introduce \textbf{Arena-T2I Hard}, a $310$-prompt stress benchmark
drawn from real arena T2I logs, with approximately $30$ decomposed yes/no
constraints per prompt spanning six categories, including text
rendering. The strongest closed-source system we evaluate reaches
$0.855$ with a $33$~pp performance gap across $11$ systems, demonstrating substantial discriminative power. Moreover, high public-arena rankings fail to predict faithfulness, confirming that
holistic Bradley-Terry (BT) preference scores prioritize aesthetics
over fine-grained prompt adherence.
We propose a \textbf{dependency-aware checklist reward} that decomposes each prompt into a DAG of yes/no questions and zeroes descendants of
failed parents, turning faithfulness into a per-constraint training signal. Combined with a BT aesthetic reward via \textbf{group-decoupled normalization (GDPO)}, which standardizes each reward within its rollout group so neither collapses, the recipe attains a strictly better faithfulness-aesthetics trade-off on SD3.5-Medium and FLUX.1-dev under MMRB2 pairwise comparisons than
every single-reward, naive weighted-sum, or 4-reward BT-ensemble
baseline. We release Arena-T2I Hard as a public stress
benchmark.
\end{abstract}
\vspace{10pt}
{\noindent
\begin{tabular}{@{}l@{\hspace{1em}}l@{}}
  \textbf{Website} & \url{https://banyuanhao.github.io/Arena-T2I-Hard-Page/} \\
  \textbf{Code and data} & \url{https://github.com/banyuanhao/Arena-T2I-Hard}
\end{tabular}\par}
\end{tcolorbox}

\section{Introduction}
\label{sec:intro}

\begin{figure}[t]
  \centering
  \includegraphics[width=0.8\linewidth]{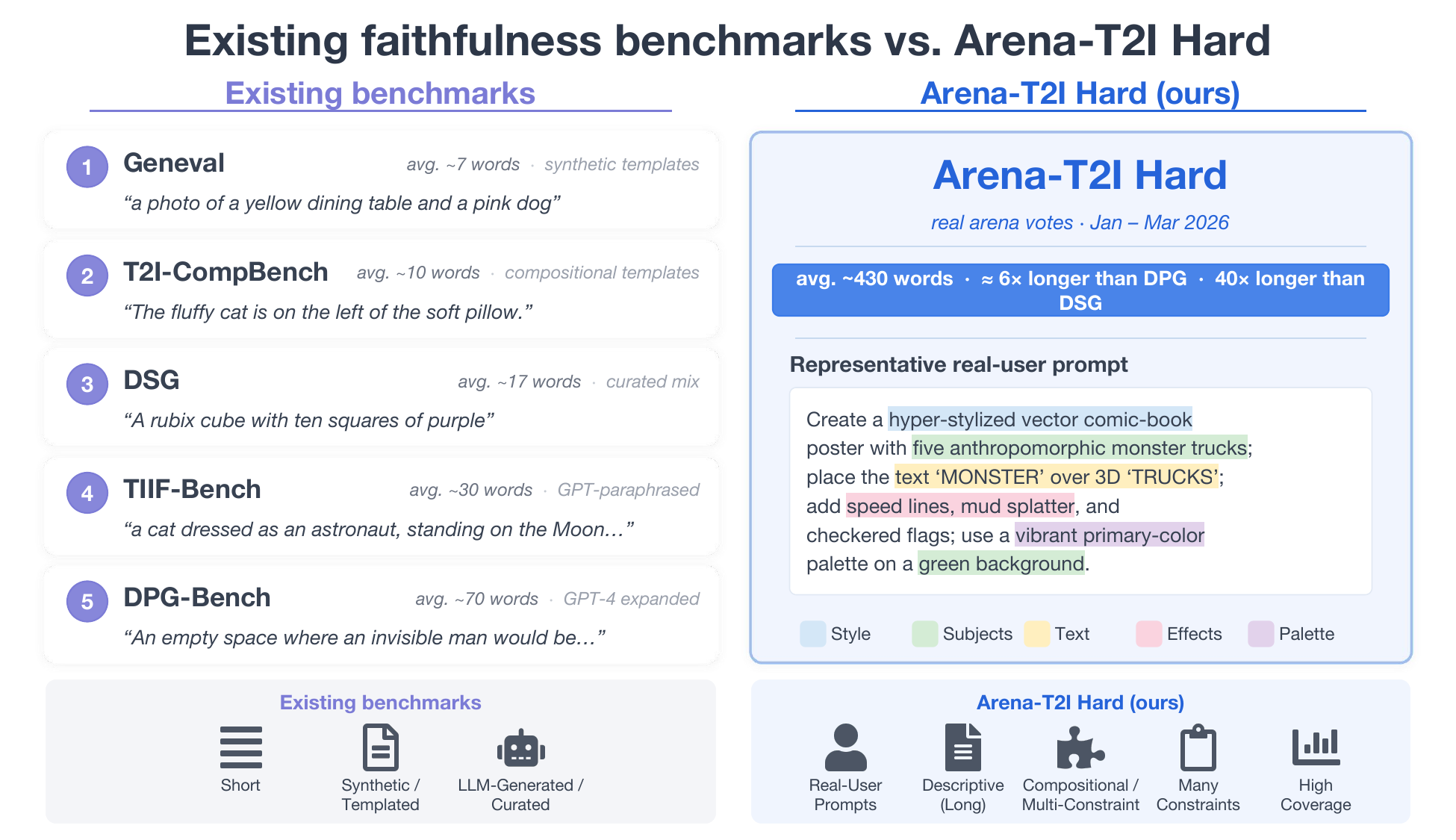}
  \caption{\textbf{One representative prompt per faithfulness
  benchmark.} Existing T2I faithfulness benchmarks rely on
  \emph{synthetic templates}, \emph{LLM
  rewriting} of short concepts, or \emph{curated short
  captions}; even the longest-form predecessor, DPG-Bench,
  averages only ${\sim}70$ words per prompt. Our \textbf{Arena-T2I
  Hard} draws prompts from real T2I-arena user votes
  and selects for compositional difficulty: prompts average
  ${\sim}430$ words and ${\sim}30$ decomposed yes/no questions.}
  \label{fig:teaser}
\end{figure}

The rapid rise of high-quality text-to-image (T2I) models has moved generative AI into practical, real-world applications~\citep{nano-banana-2,nano-banana-pro,gpt-image-1,gpt-image-2,flux2,flux1,sd3,hunyuanimage3,qwenimage2,wan2.6,recraftv4}. In these settings, faithfulness---how accurately an image follows a user's prompt---is a critical requirement. While established benchmarks like DSG~\citep{dsg}, TIFA~\citep{tifa}, and DPG~\citep{dpg} were created to measure this, they predominantly rely on simple instructions that can be easily verified. Consequently, most state-of-the-art models now achieve near-perfect scores (over 95\%) on these existing tests. However, as these models integrate into daily workflows, user requests are becoming increasingly complex and realistic. Current benchmarks struggle to measure faithfulness for these sophisticated, multi-faceted instructions. As shown in Figure~\ref{fig:teaser}, modern requests often involve intricate spatial relationships, specific styles, and complex text rendering. In these scenarios, existing benchmarks are no longer sufficient to capture the nuances of where a model succeeds or fails.

Motivated by this gap, we propose a new faithfulness benchmark: \textbf{Arena-T2I Hard}. This benchmark consists of deliberately compositional prompts sampled from the \textit{Arena Text-to-Image votes}, capturing realistic user requests ranging from 3D generation to precise text rendering. To evaluate these prompts, we propose a \textbf{dependency-aware checklist}. Instead of using a single binary score, we decompose each prompt into a directed acyclic graph (DAG) of yes/no questions. A VLM judge is then prompted to answer each question, whose responses are then aggregated to determine the final grade. If a parent question fails, its descendants are scored \textsc{no} without a further VLM call, preventing the inflated scores that flat checklists produce when an attribute question fires on the wrong object.

      \begin{table}[t]
      \centering
      \footnotesize
      \setlength{\tabcolsep}{4pt}
      \resizebox{\textwidth}{!}{%
      \begin{tabular}{rlrrrrr}
      \toprule
      \textbf{\#} & \textbf{Model}
                  & \textbf{Arena-T2I Hard $\uparrow$}
                  & \textbf{DPG-Bench $\uparrow$}
                  & \textbf{DSG $\uparrow$}
                  & \textbf{Arena rank}
                  & \textbf{Arena score} \\
      \midrule
      $1$ & \texttt{gemini-3-pro-image-preview-2k}          & $0.855$ & $0.970$ & $0.946$ & $\#3$  & $1244\pm 4$ \\
      $2$ & \texttt{grok-imagine-image-20260306}            & $0.849$ & $0.965$ & $0.934$ & $\#8$  & $1170\pm 4$ \\
      $3$ & \texttt{gpt-image-1.5-high-fidelity}            & $0.796$ & $0.970$ & $0.942$ & $\#4$  & $1242\pm 4$ \\
      $4$ & \texttt{recraft-v4}                             & $0.787$ & $0.956$ & $0.914$ & $\#29$ & $1109\pm 5$ \\
      $5$ & \texttt{wan2.6-t2i-v2}                          & $0.768$ & $0.954$ & $0.917$ & $\#21$ & $1132\pm 4$ \\
      $6$ & \texttt{gemini-2.5-flash-image} (nano-banana)   & $0.768$ & $0.954$ & $0.921$ & $\#14$ & $1152\pm 3$ \\
      $7$ & \texttt{gpt-image-1}                            & $0.722$ & $0.959$ & $0.938$ & $\#27$ & $1115\pm 3$ \\
      $8$ & \texttt{imagen-4.0-ultra-generate-001}          & $0.680$ & $0.961$ & $0.928$ & $\#17$ & $1148\pm 4$ \\
      $9$ & \texttt{imagen-4.0-generate-001}                & $0.659$ & $0.947$ & $0.913$ & $\#22$ & $1130\pm 3$ \\
      $10$ & \texttt{hunyuan-image-3.0-fal}                  & $0.609$ & $0.947$ & $0.837$ & $\#15$ & $1151\pm 3$ \\
      $11$ & \texttt{ideogram-v3-quality}                    & $0.523$ & $0.894$ & $0.855$ & $\#42$ & $1049\pm 4$ \\
      \bottomrule
      \end{tabular}%
      }
      \caption{Faithfulness yes-ratio of $11$ leading closed-source T2I
      systems on Arena-T2I Hard, DPG-Bench~\citep{dpg}, and
      DSG~\citep{dsg}, all scored by 
      gemini-3-flash. Sorted by Arena-T2I Hard faithfulness.
      DPG-Bench and DSG are saturated 
      while Arena-T2I Hard is still able to distinguish the performance between top models. Also, different rankings are observed between faithfulness and arena-leaderboard ranking, showing that the overall ranking may not be a good proxy for fine-grained prompt faithfulness.}
      \label{tab:arena-t2i-hard-leaderboard}
      \end{table}

We evaluate 11 state-of-the-art T2I models on Arena-T2I Hard (Table~\ref{tab:arena-t2i-hard-leaderboard}), revealing significant weaknesses even in top-tier models. For instance, even state-of-the-art models such as \texttt{nano-banana-pro}~\citep{nano-banana} and \texttt{grok-imagine}~\citep{grok} show non-trivial gaps in faithfulness despite their high general popularity. Crucially, our results show that a high public-arena ranking is no guarantee of high faithfulness; models like \texttt{recraft-v4}~\citep{recraftv4} rank \#29 in general preference but achieve good performance on faithfulness. 

We next study how our dependency-aware reward can be used to improve T2I models via Group Relative Policy Optimization (GRPO)~\citep{grpo}. We observe a fundamental tension in the standard RL recipe: optimizing for faithfulness alone improves adherence but degrades BT-based aesthetic rewards, while optimizing for aesthetics alone often pushes faithfulness down. To resolve this, we propose conducting GRPO with an ensemble of BT rewards and our structured faithfulness score, balanced via Group reward-Decoupled Normalization Policy Optimization~\citep{gdpo}. By training on a subset of Arena-T2I prompts with this combined reward, we demonstrate that it is possible to improve both aesthetics and faithfulness simultaneously. Based on exhaustive experiments, we conclude that \textbf{combining prompt-specific dependency-aware checklist reward and an aesthetic BT-based reward achieves the best overall performance.} This could serve as a recipe for T2I model post-training.

\paragraph{Contributions.}
\begin{compactitem}
    \item We introduce \textbf{Arena-T2I Hard}, a 310-prompt stress benchmark derived from real-world user requests to measure the faithfulness ceiling of T2I models.
    \item We develop a \textbf{dependency-aware checklist reward} that decomposes complex prompts into a structured graph of constraints, providing a more reliable training and evaluation signal than flat rubrics. 
    \item We identify a \textbf{reward--task mismatch} in current T2I RLHF, showing that standard preference rewards often optimize for aesthetics at the expense of faithfulness.
    \item We demonstrate that combining heterogeneous rewards (faithfulness + aesthetics) improves both aspects and show that the proposed post-training method can significantly improve the performance of two base models (SD3.5-Medium and FLUX.1-dev) on public image generation benchmarks.
\end{compactitem}

\section{Related Work}
\label{sec:related}

\paragraph{Reinforcement learning for text-to-image generation.}
Diffusion-DPO~\citep{ddpo} adapts direct preference optimization to
diffusion, and group-based methods such as
FlowGRPO~\citep{flowgrpo} use $K$ rollouts per prompt to estimate
group-relative advantages. T2I-R1~\citep{t2i-r1} uses semantic and token-level CoT to boost autoregressive T2I models. Concurrently, group reward-decoupled normalization~\citep{gdpo} proposes to normalize each reward first to avoid reward-scale imbalance across heterogeneous objectives.

\paragraph{T2I benchmarks.}
Geneval~\citep{geneval}, TIIF~\citep{tiif}, and
T2I-CompBench~\citep{t2icompbench} use template-based synthetic
prompts to probe object presence, attributes, and relations.
TIFA~\citep{tifa} introduces structured yes/no checklists, and
DSG~\citep{dsg} extends them with dependency-graph alignment
scoring; DPG-Bench~\citep{dpg} scales the same idea to long
GPT-augmented prompts. These benchmarks all rely on synthetic
templates, LLM rewriting, or curated scenarios, and they saturate
for top systems. Arena-T2I Hard differs in being drawn from real
arena T2I logs at $\sim$$430$ words/prompt, $\sim$$6{\times}$ longer
than DPG-Bench.

\paragraph{Reward design.}
T2I reward models are typically Bradley-Terry preference scorers
trained on pairwise human judgments~\citep{hpsv3,imagereward,pickscore};
they correlate with broad human preference but conflate aesthetics
with prompt following. UnifiedReward~\citep{unifiedreward} distills a VLM
judge into a single scalar covering multiple axes;
Rubric-RL~\citep{rubricrl} adds a free-form flat rubric without
separating aesthetics from faithfulness. We instead promote a
\emph{dependency-aware} faithfulness checklist from evaluation artifact to \emph{training} reward.
\section{Arena-T2I-Hard with Dependency-Aware Checklist Reward}
\label{sec:reward}

\subsection{Reward design}
\label{sec:reward:design}

Let $p$ be a text prompt, $x$ a generated image, $\mathcal{G}$ a frozen text-only LLM decomposer, and $\mathcal{V}$ a frozen vision--language judge. Our checklist reward is $R_{\text{chk}}(x, p) = \mathrm{Aggregate}\big(\mathcal{V}\,\big|\, x, p, G(p)\big)$, where the question graph $G(p) = \mathcal{G}(p)$ is computed once per prompt and cached; only $\mathcal{V}$ is queried in the inner loop.

\paragraph{Question graph.}

$\mathcal{G}$ maps $p$ to a directed acyclic graph $G(p) = (Q(p), E(p))$. Each node $q$ carries a yes/no question, a parent set $\mathrm{Pa}(q) \subseteq Q(p)$, and a type tag $\tau(q) \in \{\texttt{faithfulness}, \texttt{aesthetics}\}$, with edges encoding logical prerequisites (attribute and relational questions depend on the existence questions). A fixed system prompt enforces this schema (Appendix~\ref{app:impl:decompose}). In our method, the reward uses only the faithfulness subset $Q_f(p) = \{q : \tau(q) = \texttt{faithfulness}\}$.

\paragraph{Dependency-aware scoring.}
We answer questions in BFS order over $G(p)$. Roots are queried directly. For a
non-root $q$, if any parent $q' \in \mathrm{Pa}(q)$ has $y_{q'} = 0$ we set $y_q = 0$ without a VLM call; otherwise $y_q = \mathbf{1}\!\left[\mathcal{V}(x, p, q) = \textsc{yes}\right]$, with \textsc{irrelevant} answers mapped to $0$. Skipping descendants of failed parents can prevent inflated scores from attribute questions whose objects are absent and reduce VLM calls.

\paragraph{Aggregation.}
The faithfulness reward is the yes-ratio over $Q_f(p)$,
$s_f(x, p) = \frac{1}{|Q_f(p)|} \sum_{q \in Q_f(p)} y_q \in [0, 1]$, and analogously for $Q_a(p)$. For post-training, we further expose two auxiliary signals: the per-question vector $\mathbf{y}(x, p) \in \{0,1\}^{|Q_f(p)|}$ (used by the GDPO sub-modes in Section~\ref{sec:method:submodes}) and the question count $n(p) = |Q_f(p)|$.

\subsection{Reward implementation}
\label{sec:reward:impl}

By default we use Gemini-3-Pro~\citep{geminipro} as the decomposer $\mathcal{G}$ due to its strong capabilities. To verify $\mathcal{V}$'s scoring quality, we construct a $100$-prompt benchmark from real user prompts sampled from Arena Text-to-Image votes. We generate one
image per prompt with SD3-Medium~\cite{sd3}, decompose each prompt into yes/no
questions ($1{,}810$ total, $\sim$$18$ per prompt), and \emph{ask a human
annotator to label every (image, question) pair}. We compare two design choices: the VLM judge base model and the query mode. Full system prompts and
per-judge $/$ per-mode metrics are deferred to Appendix~\ref{app:impl:decompose}, Appendix~\ref{app:impl:pool}, and Appendix~\ref{app:impl:judge-eval}.

\paragraph{VLM judge base model.}
We compare four candidates under oneshot mode. Gemini-3-flash-preview~\citep{geminiflash} is the most accurate at $93.0\%$, while Qwen3.5-27B~\citep{qwen35} is slightly worse at $91.9\%$ but is easier to serve at training scale. We therefore choose Qwen3.5-27B as our VLM judge $\mathcal{V}$. Please refer to Appendix~\ref{app:impl:judge-eval} and Figure~\ref{fig:prompt48-judges} for more details.

\paragraph{Query mode.}
Two query modes are supported. \textbf{Oneshot} sends all of $Q(p)$ in a single
VLM call returning a JSON array of yes/no answers; \textbf{individual} sends one
call per question. The two modes give very similar faithfulness scores. On the $100$-prompt benchmark, their per-image yes-ratios are highly correlated across all judges
(Pearson $r\geq 0.89$), with a $93.8\%$ average per-question agreement. Oneshot is also slightly more accurate
than individual querying for all judge families. We use oneshot by default for its
 lower API cost without scoring-quality loss.

Based on the above analysis, we choose Qwen3.5-27B in oneshot mode as
$\mathcal{V}$ and Gemini-3-Pro~\citep{geminipro} as $\mathcal{G}$. 

\subsection{Arena-T2I Hard Benchmark construction}
\label{sec:arena-t2i}

\paragraph{Arena Text-to-Image prompt pool.} All prompts in this paper are sampled from a public text-to-image arena
leaderboard, where users submit prompts and vote on the resulting images,
so the underlying distribution reflects the prompts that real users
actually issue to T2I systems. We select the user prompts spanning from Jan 2026 to March 2026. We apply NSFW, PII, and invalid-prompt (non-T2I user requests, e.g., image-edit prompts that require an additional input image) filtering to the raw submissions and form the arena prompt pool. We further manually check each prompt to avoid any legal issues. 

\paragraph{Arena-T2I Hard.}
To stress-test faithfulness specifically and give a benchmark that
remains discriminative even for the strongest T2I systems, we
construct \textbf{Arena-T2I Hard}, a deliberately hard
benchmark of \textbf{310 prompts} drawn from the arena prompt pool and
selected for compositional difficulty: long, multi-entity, with explicit
attributes, spatial relations, counts, and stylistic constraints stacked
on top of one another. Each prompt is decomposed by the Gemini-3-Pro
pipeline and carries on average $\sim$30 yes/no faithfulness questions,
with dependency-graph depth up to $5$ and as many as $10$ direct
dependencies on a single attribute or relation question. The $310$
prompts are roughly balanced across six visual-style categories
(\emph{art}, \emph{3d\_modeling}, \emph{cartoon}, \emph{photorealistic},
\emph{portraits}, \emph{commercial\_design}); the per-category breakdown
is reported in Appendix~\ref{app:bench:hard}.

\paragraph{RL training and testing datasets.}
To construct an easier dataset for RL training on weaker open-source models, we sample two disjoint subsets \emph{uniformly at random} 
from the arena prompt pool: a
\textbf{10k training set} used for all RL runs, and a \textbf{1k test set}
used as the held-out evaluation set for all evaluation results.
Because both subsets are i.i.d.\ samples from the arena distribution, win rates on the test set serve as an unbiased estimate of the underlying real-world prompt distribution.

\subsection{Leaderboard computing and discussions}
\label{sec:leaderboard}

We benchmark $11$ leading closed-source T2I systems on Arena-T2I
Hard; for comparison we score every system on
DPG-Bench~\citep{dpg} and DSG~\citep{dsg} under the same judge.
Table~\ref{tab:arena-t2i-hard-leaderboard} reports the leaderboard.

\paragraph{Headroom remains and only Arena-T2I Hard discriminates.}
The strongest model, \texttt{gemini-3-pro-image-preview-2k}, reaches
$0.855$---$14$~pp from ceiling and the top of an $11$-system spread
of $33$~pp down to $0.523$. The same systems on DPG-Bench and DSG
are saturated ($0.89$--$0.97$ and $0.84$--$0.95$).

\paragraph{Public arena rank is a weak proxy for faithfulness.}
Several public top-$15$ systems drop sharply
(\texttt{hunyuan-image-3.0} \#$15{\to}10$th at $0.609$; nano-banana
\#$14{\to}6$th at $0.768$), while \texttt{recraft-v4} climbs from
\#$29$ to $4$th at $0.787$. The pattern indicates
that humans sometimes prioritize aesthetics over fine-grained prompt adherence.

\paragraph{Decomposer robustness.}
To assess the robustness of the decomposer, we re-decompose all $310$ prompts
using a second decomposer (GPT-5.4) and rescore them with the same judge. The
resulting per-model mean yes-rate vectors are highly correlated with the
original scores, with Pearson correlation $0.991$ and Spearman correlation
$0.982$. Although all systems' scores decrease uniformly by $0.05$--$0.09$, no
system changes by more than one rank. See Appendix~\ref{app:bench:hard} and
Table~\ref{tab:decomposer-robustness} for details.
\section{Post Training by Combining Faithfulness with Aesthetics}
\label{sec:method}


In this section, we study whether the
proposed dependency-aware checklist can serve not only as an evaluation
tool, but also as a training signal for improving prompt following. To this
end, we instantiate the checklist score as a reward in Flow-GRPO~\citep{flowgrpo} and study
how it interacts with standard preference-based rewards. Our analysis
reveals a central challenge: optimizing a single reward improves the
corresponding metric but often fails to transfer across axes, and may even
degrade other aspects of generation quality. This motivates a combined
post-training objective that explicitly balances faithfulness and aesthetics.

\subsection{Post-training with single scalar rewards}

\label{sec:method:collapse}
We use GRPO~\cite{grpo}, which is a standard practice in T2I RL. For
each prompt $p_i$ we sample $K$ rollouts $\{x_{i,j}\}_{j=1}^K$, compute
$M$ scalar rewards $r_k(i,j)$, and form the group-relative advantage of
a weighted sum:
\begin{equation}
  r(i,j) = \sum_{k=1}^{M} w_k \, r_k(i,j),
  \qquad
  A(i,j) = \frac{r(i,j) - \mu_i}{\sigma_i + \varepsilon},
  \label{eq:grpo}
\end{equation}
with $\mu_i, \sigma_i$ the group mean and standard deviation. 

\paragraph{Single reward design.} For reward choice, we try two open-source BT rewards trained from human preference data as well as our faithfulness reward.
\textit{PickScore} is a CLIP-H/14-based BT reward trained on ${\sim}500$K Pick-a-Pic pairwise human preferences~\citep{pickscore}. Since these preferences are open-ended and the CLIP dual-tower architecture has limited compositional reasoning, PickScore mainly captures aesthetic appeal rather than fine-grained prompt fidelity. \textit{HPSv3} is a stronger Qwen2.5-VL-7B-based BT reward trained on the ${\sim}1$M-pair HPDv3 dataset, whose annotations cover both aesthetics and prompt faithfulness~\citep{qwen2_5_vl,hpsv3}. \textit{Our faithfulness reward} use the dependency-aware checklist as described in Sec~\ref{sec:reward}. 

\begin{figure}[h]
  \centering
  \includegraphics[width=0.85\linewidth]{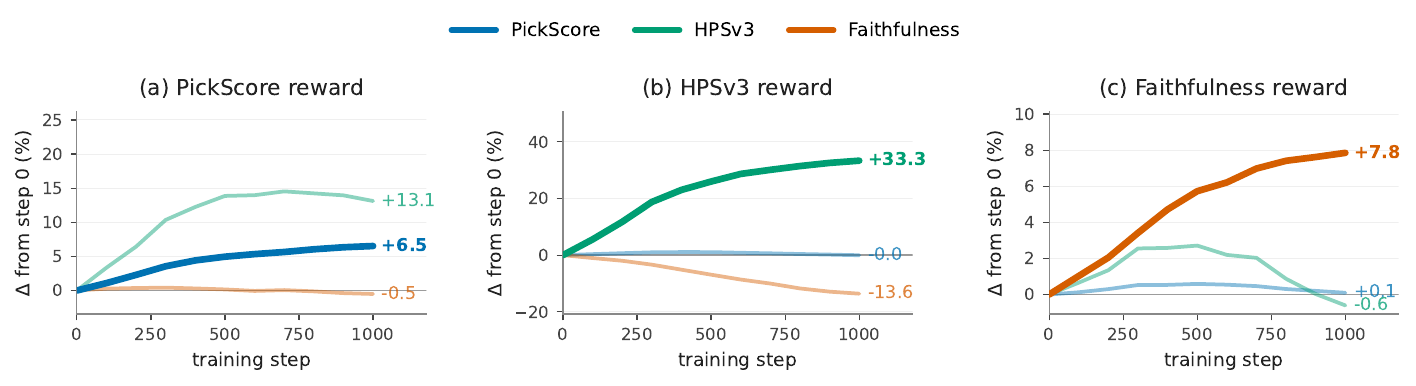}
  \caption{Eval-reward dynamics for three single-reward fine-tunes on SD3.5-M,
  truncated at training step 1,000. In each panel the optimized reward is
  shown in full opacity, eval-only rewards in lighter ink. We observe that BT-style training
  drives faithfulness flat or below baseline and faithfulness-only training
  does not lift the BT rewards.}
  \label{fig:training-curves}
\end{figure}

\paragraph{Single-reward training does not transfer across axes.} We finetune a SD3.5-M~\citep{sd3} on 10k training set for 1,000 training steps with three single rewards separately. We evaluate the ckpts on 1k hold out testing prompts. Figure~\ref{fig:training-curves} plots the relative change of three eval rewards over 1,000 training steps. More details can be found in Appendix~\ref{app:hparams}. 
\emph{HPSv3-only} climbs much more sharply ($+33\%$) and drives faithfulness down to roughly $-14\%$.
\emph{Faithfulness-only}, conversely, climbs its own reward $+8\%$ but does
not lift PickScore or HPSv3.
This reveals a clear reward--axis mismatch. (1) Optimizing a single reward does not necessarily transfer to other evaluation axes, and gains in aesthetics can come at the cost of faithfulness. (2) Strong Bradley--Terry preference models, despite being trained on large-scale human preference data, may still provide poor RL training signals for improving prompt faithfulness.

\subsection{Combining the two rewards via GDPO lifts both eval signals}
\paragraph{Pitfalls in current reward ensembling methods.} A direct way is to use the weighted sum of the two rewards. However, when the
components of $\mathbf{r}$ have very different within-group scales,
$\sigma_i$ is dominated by the highest-variance term, and it is extremely hard to tune the weighting parameter of the combining reward.
In our setting, the BT preference reward has significantly different dynamics from the faithfulness $0/1$ checklist reward, so naive Eq.~\eqref{eq:grpo} often does not perform well.

\paragraph{Group-decoupled normalization policy optimization.}
\label{sec:method:gdpo}
To resolve this issue, we adopt group-decoupled normalization (GDPO)~\citep{gdpo}, which normalizes each reward within its rollout group \emph{before} combining:
\begin{equation}
  A_k(i,j) = \frac{r_k(i,j) - \mu_{k,i}}{\sigma_{k,i} + \varepsilon},
  \qquad
  A(i,j) = \sum_{k=1}^{M} w_k \, A_k(i,j),
  \label{eq:gdpo}
\end{equation}
where $\mu_{k,i}, \sigma_{k,i}$ are the per-prompt statistics of the
$k$-th reward across the $K$ rollouts. Each component contributes an
advantage of unit scale by construction. It stabilizes training and setting $\{w_k\}$ can always achieve good results. Please refer to Appendix~\ref{app:results:gdpo} for more details.

We fine-tune FLUX.1-dev~\cite{flux1} for 1250 steps on the same 10k-prompt training set. Experimental details are provided in Appendix~\ref{app:hparams}. Figure~\ref{fig:training-curves-combined} reproduces the reward trade-off on FLUX.1-dev and adds a third panel for our combining faithfulness and pickscore under GDPO. The single-reward runs, shown in panels (a) and (b), mirror the SD3 results: \emph{PickScore-only} reduces faithfulness by $1.1\%$, while \emph{Faithfulness-only} reduces PickScore by $2.5\%$. In contrast, the combined run in panel (c) improves both evaluation rewards simultaneously, increasing faithfulness by $10.5\%$ and PickScore by $3.5\%$. GDPO therefore avoids reward collapse; moreover, its faithfulness gain exceeds that of the dedicated faithfulness-only run ($+6.8\%$) under the same step budget. Additional comparisons between GRPO and GDPO are provided in Appendix~\ref{app:results:gdpo}, which demonstrates GDPO are more effective than GRPO.

\begin{figure}[h]
  \centering
  \includegraphics[width=0.85\linewidth]{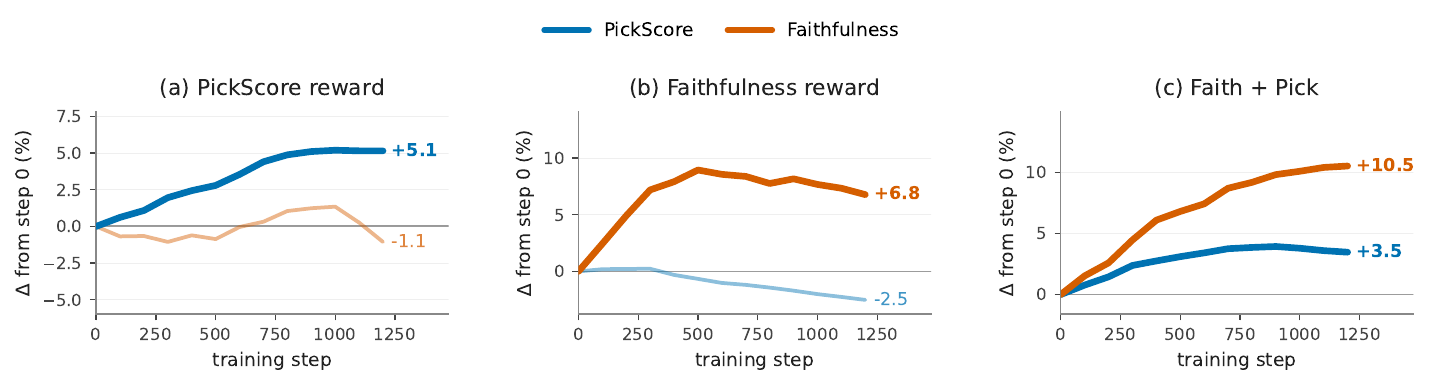}
  \caption{Eval-reward dynamics on FLUX.1-dev for two single-reward
  fine-tunes and a combined Faith\,+\,Pick run trained under GDPO,
  truncated at step 1,250. In each panel the optimized reward(s) are shown in
  full opacity. Numbers at the right edge are the final $\Delta$
  from step 0 in \%. We observe that single-reward training degrades the cross-axis reward (panels a, b) while GDPO lifts both (panel c).}
  \label{fig:training-curves-combined}
\end{figure}

\section{Main Experiments}
\label{sec:experiments}
\label{sec:results}
The previous section shows that combining aesthetic and faithfulness rewards can mitigate the reward--task mismatch: optimizing preference rewards alone tends to improve aesthetics while sacrificing faithfulness. In this section, we further analyze reward design and investigate which combination strategy best supports joint improvement in post-training.

\paragraph{Evaluation protocol.} Existing T2I RL works typically report progress on the same reward family used for training, or on benchmarks targeting the same dimension as the reward. Such reward-aligned evaluation mainly measures how well the policy fits its training signal, rather than whether it improves along other dimensions. To better simulate real-world scenario, we evaluate all main results on the held-out 1k arena-distribution test set from Section~\ref{sec:arena-t2i} using the \emph{MMRB2 pairwise protocol}~\citep{mmrb2}. For each pair of fine-tuned models, we generate one image per prompt from each model and ask an independent Gemini-3-flash judge to compare them under a fixed MMRB2 rubric covering faithfulness, text--image alignment, text rendering, and overall image quality. The judge produces a single preference verdict for each pair, and we aggregate these verdicts into a net win rate across the 1,000 prompts. Each pair is judged twice with the order swapped to reduce position bias. Additional details on the evaluation protocol and the MMRB2 rubric are provided in Appendix~\ref{app:eval}.

\paragraph{Training settings.}
We follow the GRPO/GDPO training protocol of FlowGRPO~\citep{flowgrpo} and evaluate on two open text-to-image backbones: \textbf{Stable Diffusion 3.5-Medium} (SD3.5-M)~\citep{sd3} and \textbf{FLUX.1-dev}~\citep{flux1}. Full hyperparameters are in Appendix~\ref{app:hparams}.


\paragraph{Single-reward baselines.} We adopt four publicly available preference rewards, each used alone with weight $1$: PickScore~\citep{pickscore}, HPSv3~\citep{hpsv3}, ImageReward~\citep{imagereward}, and UnifiedReward-2.0~\citep{unifiedreward}. These rewards differ in backbone and supervision: PickScore is a CLIP-based BT reward trained on Pick-a-Pic preferences, HPSv3 is a Qwen2.5-VL-based BT reward trained on HPDv3, ImageReward uses a BLIP backbone with expert-annotated rankings and is trained in a BT way, and UnifiedReward-2.0 is a VLM-based scalar judge designed to cover both aesthetics and prompt following. Please refer to Appendix~\ref{app:training_details:rms} for more details.


\paragraph{Ensembling rewards.} Our \emph{Faith\,$+$\,Pick} pairs PickScore as the aesthetic signal with our checklist as the faithfulness signal, and trains under GDPO. We use a flat weight of $1$ on each reward. We also include a 4-reward
\emph{ensemble} that linearly combines the four single-reward models
above at scale-balanced weights. Thus, any improvement of our Faith\,$+$\,Pick recipe over this ensemble reflects the value of the checklist-based faithfulness signal and careful reward design, rather than simply the benefit of combining more rewards.

\paragraph{Iterative SFT baseline.}
We also compare against a DreamSync-style iterative SFT baseline~\citep{dreamsync}, which alternates between generating candidate images, filtering them with reward scores, and fine-tuning on the retained prompt--image pairs. At each iteration, the current policy samples $8$ candidates per prompt on the same 10k training set, scores them with PickScore and our checklist reward, and retains one reward-selected target per prompt when available. The retained pairs are used for a 1000-step flow-matching LoRA fine-tune, with each iteration initialized from the previous LoRA checkpoint. We run three iterations on each backbone. Full generation, filtering, and fine-tuning details are provided in Appendix~\ref{app:dreamsync}.

\label{sec:results:main}
\label{sec:results:winrate}
\label{sec:results:axis}
\subsection{Is our combined reward better than single or naive reward mixing?}

\begin{figure}[t]
  \centering
  \begin{subfigure}{0.44\linewidth}
    \centering
    \includegraphics[width=\linewidth]{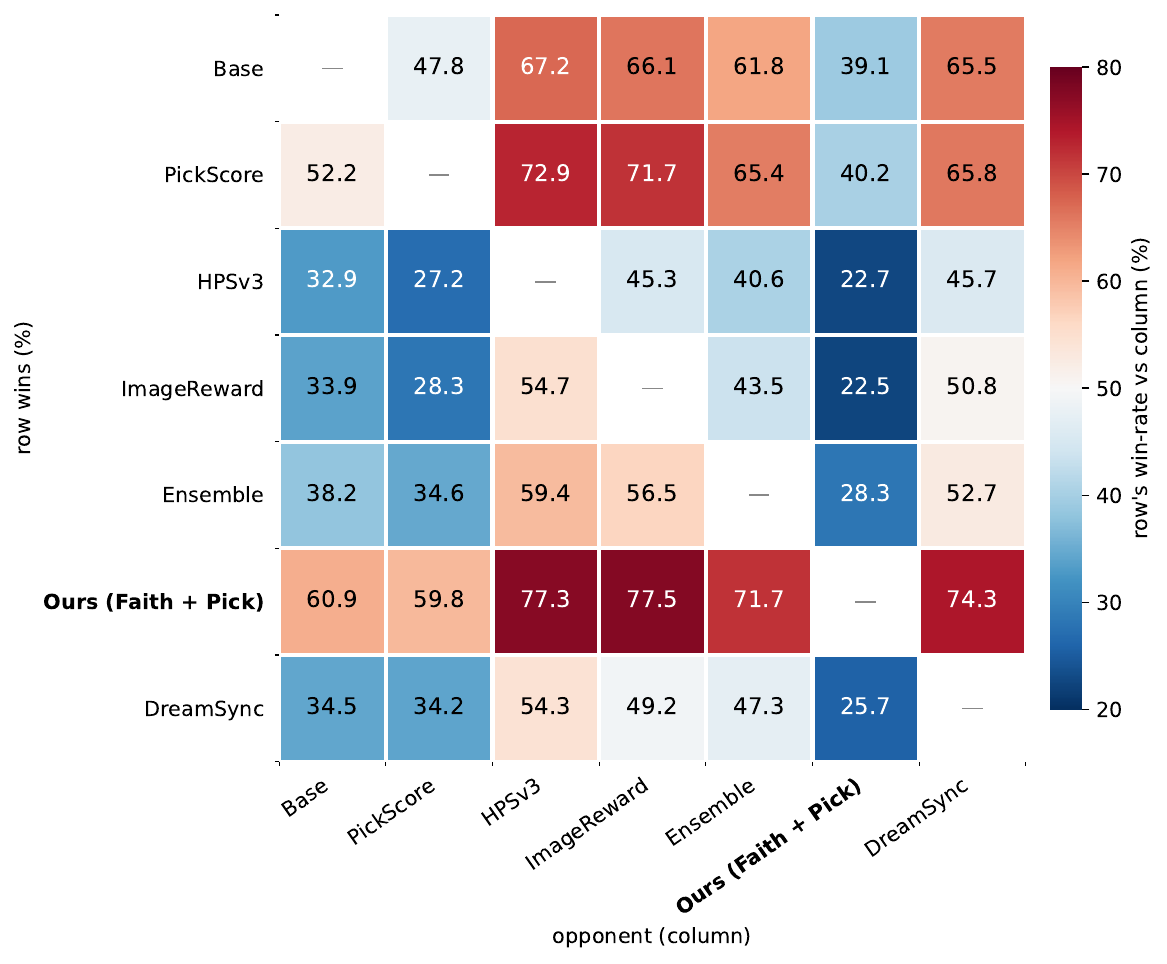}
    \caption{SD3.5-M}
    \label{fig:matrix-sd3-baseline}
  \end{subfigure}
  \hfill
  \begin{subfigure}{0.44\linewidth}
    \centering
    \includegraphics[width=\linewidth]{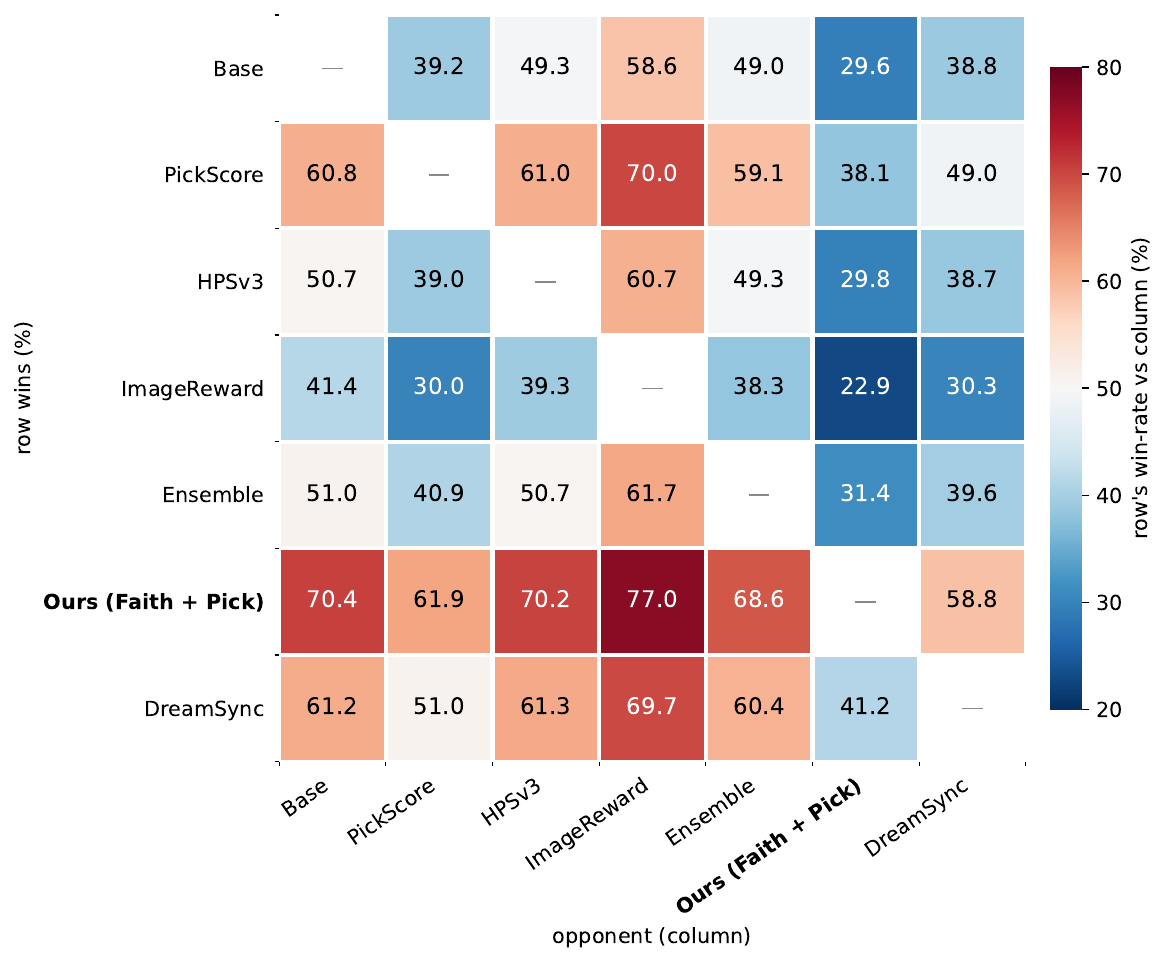}
    \caption{FLUX.1-dev}
    \label{fig:matrix-flux-baseline}
  \end{subfigure}
  \caption{Pairwise net win-rate matrices on the 1k test set, judged
  by Gemini-3-flash under the MMRB2 rubric. Each cell $(i,j)$ is row
  $i$'s win rate against column $j$ at \texttt{image\_idx}${=}0$, with
  each prompt judged twice under swapped order. Ours (Faith\,$+$\,Pick GDPO)
  wins against every single-reward and iterative-SFT baseline on both
  backbones; cross-seed stability of these rankings is reported in
  Appendix~\ref{app:results:rowmean}.}
  \label{fig:matrix-baseline}
\end{figure}

\paragraph{Pairwise VLM judge win-rate results.}
Figure~\ref{fig:matrix-baseline} compares all baselines using the MMRB2 pairwise protocol on the 1k test set. Across both SD3.5-M and FLUX.1-dev, our combined Faith\,$+$\,Pick GDPO run achieves the strongest row-mean win rate and beats every single-reward baseline as well as the iterative SFT baseline. The ranking is stable across three random-seed rounds with cross-round standard deviations below $1.2$~pp; see Appendix~\ref{app:results:rowmean} for the row-mean bar charts.

\paragraph{Single BT rewards: marginal or detrimental effects.}
On SD3.5-M, training on PickScore alone yields only a marginal preference
improvement over the base backbone ($52.2\%$), HPSv3 is actively
\emph{worse} than base ($32.9\%$), and ImageReward is well below base
($32.8\%$). On FLUX.1-dev PickScore alone is more competitive ($60.8\%$ over base),
while HPSv3 is roughly tied with base ($50.7\%$). This shows that BT preference rewards optimized in isolation range from no-better-than-base to substantially worse on the held-out MMRB2 verdict.

\paragraph{Linear ensembling of multiple rewards does not help.}
On SD3.5-M, the 4-reward preference ensemble (PickScore $+$ HPSv3 $+$
ImageReward $+$ UnifiedReward-2.0 at scale-balanced weights) wins only
$38.2\%$ against base. On FLUX, the ensemble is at parity with base ($51.0\%$).
Linearly combining multiple open-source top rewards does
not unlock a qualitatively new signal; it amplifies the aesthetic bias
shared across the four.

\paragraph{Faith+Pick is the strongest row on both backbones.}
Our Faith\,$+$\,Pick run wins every cell of its row against every BT
single-reward baseline and against the 4-reward ensemble. Pairing the dependency-aware faithfulness checklist with
PickScore through GDPO therefore yields a better
faithfulness--aesthetics trade-off than every single-reward,
ensemble, or naive weighted-sum baseline on both backbones, indicating successful T2I post-training needs careful reward design, not just more rewards.

\paragraph{GDPO outperforms linear reward summation.}
GDPO is more effective than directly summing the faithfulness and BT aesthetic rewards with fixed linear weights, achieving a pairwise win rate of $51.3\%$. See Figure~\ref{fig:matrix-gdpo-medium} and Appendix~\ref{app:results:gdpo} for experimental details.

\paragraph{Our method substantially improves base-model faithfulness.}
Table~\ref{tab:arena-t2i-hard-trained} and Appendix~\ref{app:bench:hard} shows that our method achieves the best results on both Arena-T2I Hard and DPG-Bench. On SD3.5-M, it surpasses the second-best method by $5.2\%$ on Arena-T2I Hard and $3.7\%$ on DPG-Bench, demonstrating that our method improves faithfulness and generalizes to other evaluation settings well.

\paragraph{Human-judge validation.}
The Matrix~1 SD3.5-M results above are judged by Gemini-3-Flash under the MMRB2 rubric. To check whether this judge is aligned with human preferences, we ran a small human study using \emph{Faith\,$+$\,Pick} as the anchor against the six other Matrix~1 models. We sampled $300$ stratified prompts per cell, with an additional $300$-prompt mini-study for the closest comparison. In total, we collected $1{,}899$ pairwise A/B votes with randomized side assignment and the VLM verdict hidden. Faith\,$+$\,Pick wins $64.1\%$ overall ($1{,}218/1{,}899$ pairs) and outperforms every individual opponent: ImageReward ($95.5\%$), Ensemble ($70.2\%$), DreamSync ($64.9\%$), HPSv3 ($63.0\%$), PickScore ($56.2\%$), and Base-SD3 ($55.9\%$). The human votes agree with Gemini-3-Flash on the direction of the Faith\,$+$\,Pick advantage in every cell. Full per-cell counts and side-by-side comparisons with the VLM judge are provided in Appendix~\ref{app:results:human-study}, Table~\ref{tab:human-study}.

\subsection{What Makes Effective Faithfulness Rewards?}
\label{sec:ablations}
\label{sec:ablations:questions}

In this section, we ablate different question-generation strategies and study where checklist rewards are most effective. Here are the methods compared in this ablation study:

\begin{compactitem}
    \item \textbf{Default (\emph{Faith\,$+$\,Pick}).}
    PickScore paired with the prompt-specific, dependency-aware faithfulness checklist under GDPO.

    \item \textbf{\emph{Ignore-dep.}}
    This variant removes BFS gating at scoring time: all questions are treated as roots, so child attributes, relations, and counts can score \textsc{yes} even when their parent existence question fails. See Appendix~\ref{app:ablations} for details.

    \item \textbf{\emph{Generic}.}
    This variant pairs PickScore with a fixed 15-question, prompt-agnostic checklist covering general faithfulness criteria, such as overall prompt fidelity and key-object presence. See Appendix~\ref{app:ablations} for the full rule list.

    \item \textbf{\emph{Faith\,$+$\,Aesth}.}
    This variant replaces PickScore with a second checklist of aesthetic-quality questions from the aesthetic decomposition pass (Appendix~\ref{app:impl:decompose}). We concatenate the faithfulness and aesthetic subgraphs and use their combined yes-ratio as the only reward. This tests whether checklist-style rewards can cover both axes, or whether aesthetics benefits more from a BT preference signal.

    \item \textbf{\emph{RubricRL}.}
    This variant follows RubricRL~\citep{rubricrl}: Gemini-3-Pro decomposes each prompt into a flat, prompt-adaptive rubric of roughly 10 yes/no items spanning faithfulness and aesthetics. All items are treated as root questions, and the reward is the unweighted yes-ratio over the rubric. The decomposer prompt is provided in Appendix~\ref{app:ablations}.
\end{compactitem}

Figure~\ref{fig:matrix-rules-ablation} ablates the structure of the
checklist reward on SD3.5-M at ckpt-1000. All variants share the
training-side hyperparameters of Section~\ref{sec:experiments}, run
under GDPO with a flat weight of $1$ on each reward component, and
are evaluated on the same 1k held-out test set.

\begin{wrapfigure}{r}{0.36\linewidth}
  \centering
  \vspace{-\intextsep}
  \includegraphics[width=\linewidth]{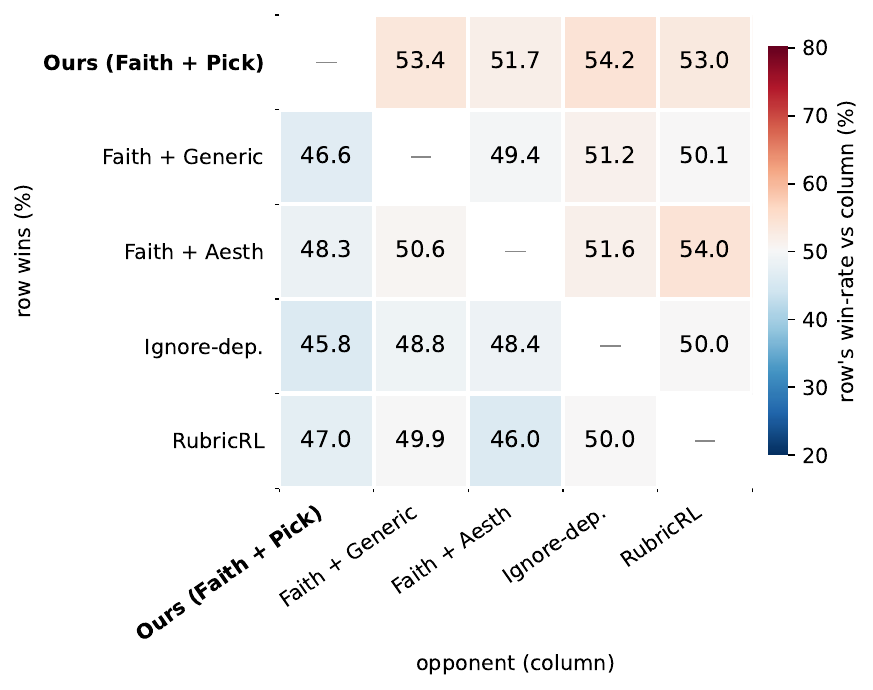}
  \caption{Question-style ablation on SD3.5-M.}
  \label{fig:matrix-rules-ablation}
\end{wrapfigure}
The ablation validates three design choices. First, the \textbf{BFS dependency walk is important}: removing it (\emph{Ignore-dep.}) causes the largest drop, with only a $45.8\%$ win rate against our full method, corresponding to a $4.2$-point swing. Its benefit also appears within the ablation block: \emph{Faith\,$+$\,Aesth}, which can be viewed as \emph{RubricRL} equipped with the same dependency walk over the mixed faithfulness-and-aesthetics question set, beats \emph{RubricRL} with a $54.0\%$ win rate. This suggests that the dependency walk is not just a scoring heuristic, but a way to encode which questions should matter most: if an object is missing, its attributes and relations should not be rewarded independently.
Second, \textbf{prompt-specific decomposition helps}: generic rules reduce performance by $3.4$ points because they miss prompt-level compositional structure. 
Finally, an \textbf{aesthetic checklist is less effective than a BT aesthetics reward trained from preference data}. It gives a $1.7$-point drop while requiring substantially more computation, roughly $10\times$ in our setup. Overall, these results support our design choice: use a prompt-specific, dependency-aware checklist for faithfulness, and pair it with a BT preference reward for aesthetics, so that the two rewards improve complementary axes while reducing the faithfulness--aesthetics trade-off.
\section{Conclusion}
\label{sec:conclusion}

We study how to secure faithfulness in modern text-to-image generative models. We introduce Arena-T2I Hard, a stress-test benchmark for top T2I models, and a dependency-aware checklist score that measures faithfulness through structured, verifiable prompt constraints. We then use this checklist score as an RL reward for post-training. Across extensive experiments, we find that effective T2I RL requires matching the reward structure to the dimension being optimized. Prompt faithfulness is best served by prompt-specific, dependency-aware checklist rewards, whereas aesthetics is better captured by BT preference rewards trained on human comparisons. Combining these two complementary reward structures improves both faithfulness and aesthetics, while reducing the trade-off between them.


\bibliographystyle{iclr2027_conference}
\bibliography{references}


\clearpage
\appendix
\pdfbookmark[0]{Appendix}{appendix}
\section*{Appendix}

\section{Dataset Construction}
\label{app:bench}

\subsection{Source pool and filtering}
\label{app:bench:pool}

All prompts in the paper---training, test, and Arena-T2I Hard---are sampled from
a public T2I arena leaderboard where users submit prompts and vote on the
resulting images. We apply three filters to the raw user submissions:
(i)~\textbf{NSFW filtering} to remove unsafe content; (ii)~\textbf{PII
filtering} to remove prompts containing personally identifying information;
and (iii)~\textbf{invalid-prompt filtering} to remove submissions that are not
genuine T2I user requests (e.g.\ chat snippets, debug strings, empty prompts).
The remaining prompts are deduplicated by exact match and filtered for
length and language (English). The three subsets used in this paper are
sampled from this filtered pool.

\subsection{Decomposition}
\label{app:bench:decomp}

Each prompt is decomposed into faithfulness and aesthetics question sets using the
Gemini-3-Pro pipeline of Appendix~\ref{app:impl:decompose}. Decomposition is cached and
reused across all training and evaluation runs. After decomposition, each prompt has on
average $18.5$ faithfulness questions and $10.6$ aesthetics questions; the
dependency-graph depth on the faithfulness subset has mean $1.7$ and maximum $7$
across the training set (Appendix~\ref{app:training}).

\subsection{Training set}
\label{app:training}

We draw $10{,}000$ prompts uniformly at random from the filtered source pool of
Appendix~\ref{app:bench:pool} as the RL training set. After offline decomposition,
the set contains $182{,}230$ faithfulness questions and $104{,}900$ aesthetics
questions ($287{,}130$ total). $9{,}837$ of the $10{,}000$ prompts ($98.4\%$) are
decomposed into both faithfulness and aesthetics questions; $9$ prompts have only
faithfulness questions; $154$ prompts produced no parseable questions and are
dropped from the training set, leaving $9{,}846$ usable prompts.

\begin{figure}[t]
  \centering
  \includegraphics[width=\linewidth]{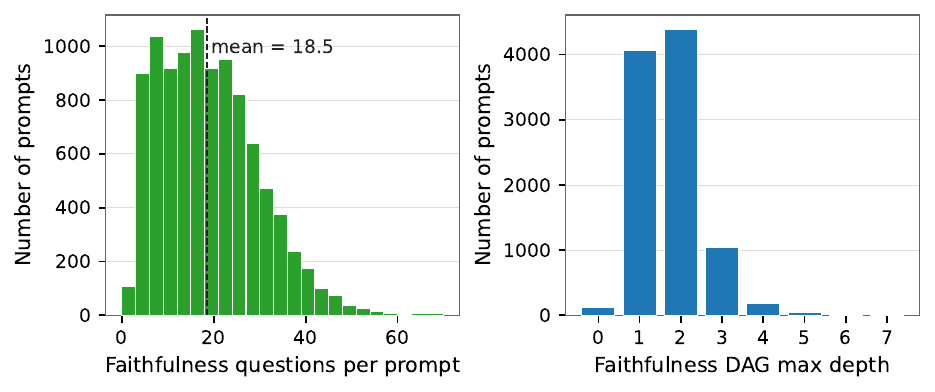}
  \caption{Distributions over the $9{,}846$ usable training prompts.
  \textbf{Left}: faithfulness questions per prompt; long-tailed with
  median $17$ and a small fraction reaching $50$+. \textbf{Right}:
  maximum DAG depth of the faithfulness subset; most prompts have
  depth $1$ or $2$, with a tail up to $7$.}
  \label{fig:training-stats}
\end{figure}

Table~\ref{tab:training-stats} summarises per-prompt statistics for the
faithfulness subset. Two features of the distribution are worth flagging.
First, the question count is heavy-tailed (Figure~\ref{fig:training-stats},
left): the median prompt has $17$ faithfulness questions, but the $99$th
percentile is $47$ and the maximum is $70$, which motivates the
NaN-padded fixed-width tensor of $\textsc{MaxQ}=128$ described in
Section~\ref{sec:reward:design}. Second, dependency-graph depth is small
on average (median $2$, mean $1.7$) but reaches up to $7$
(Figure~\ref{fig:training-stats}, right), so the BFS dependency walk has
non-trivial work to do on the deeper prompts; the maximum number of
\emph{direct} parents of a single question is $12$, and the maximum
fan-in (most-cited single question) is $52$.

\begin{table}[t]
  \centering
  \small
  \begin{tabular}{lrrrrr}
  \toprule
  & \textbf{mean} & \textbf{median} & \textbf{std} & \textbf{p95} & \textbf{max} \\
  \midrule
  Faithfulness questions / prompt & $18.5$ & $17$ & $10.6$ & $38$ & $70$ \\
  Aesthetics questions / prompt   & $10.6$ & $10$ & $\phantom{0}2.8$ & $15$ & $32$ \\
  Root (parent-less) faith.\ Q / prompt & $\phantom{0}6.9$ & $\phantom{0}6$ & $\phantom{0}4.2$ & $15$ & $42$ \\
  Faith.\ DAG max depth          & $\phantom{0}1.7$ & $\phantom{0}2$ & $\phantom{0}0.8$ & $\phantom{0}3$ & $\phantom{0}7$ \\
  Faith.\ max in-degree / prompt   & $\phantom{0}6.3$ & $\phantom{0}5$ & $\phantom{0}4.5$ & $15$ & $52$ \\
  Faith.\ max parents per question & $\phantom{0}1.8$ & $\phantom{0}2$ & $\phantom{0}0.7$ & $\phantom{0}3$ & $12$ \\
  Prompt length (characters)       & $602$ & $393$ & $852$ & $1750$ & $17467$ \\
  \bottomrule
  \end{tabular}
  \caption{Per-prompt statistics for the $9{,}846$ usable prompts in the
  $10$k training set. \emph{In-degree} of a question is the number of
  child questions that depend on it. \emph{Max depth} is the longest
  root-to-leaf path in the faithfulness DAG.}
  \label{tab:training-stats}
\end{table}

\subsection{Test set}
\label{app:bench:test}

We additionally sample $1{,}000$ prompts uniformly at random from the same
filtered pool, disjoint from the training set, as the held-out test set used
for all main MMRB2 win-rate results. Because both subsets are i.i.d.\ samples
from the same distribution, win rates on the test set are an unbiased
estimate of the win rate on the underlying real-world prompt distribution.
Each test prompt is decomposed by the same Gemini-3-Pro pipeline; per-prompt
question counts and DAG-depth statistics match those of the training set
within sampling noise (Table~\ref{tab:training-stats}).

\subsection{Arena-T2I Hard}
\label{app:bench:hard}

\paragraph{Selection protocol.}
\textbf{Arena-T2I Hard} is a deliberately hard benchmark of $310$
prompts drawn from the same filtered pool used for the training and
test sets (Appendix~\ref{app:bench:pool}). We manually inspect prompts
from this pool and select those with the longest prompt text and the
largest number of yes/no questions produced by the Gemini-3-Pro
decomposition (Appendix~\ref{app:impl:decompose}). Concretely, we
favour prompts that are simultaneously extremely long (so that the
prompt itself stacks many independent visual constraints) and decompose
into many faithfulness questions (so that the resulting checklist
exercises the dependency walk on a wide DAG).

\begin{figure}[t]
  \centering
  \begin{subfigure}{0.49\linewidth}
    \centering
    \begin{minipage}{\linewidth}
      \scriptsize\itshape
      \textbf{\upshape Prompt} (excerpt; \textbf{\upshape category:}
      portraits): A single, centered, full-body portrait of a young
      woman in her early $20$s with very long waist-length wavy deep
      burgundy hair, gray/amber eyes featuring razor-sharp
      \textbf{\upshape Polycoria} (two distinct circular pupils per
      iris, four total), a `\upshape my body your choice' script
      tattoo on her right clavicle, a silver cross earring in her
      left ear, lip and nose rings, a ribbed black crop top, high-waist
      distressed denim jeans, white leather sneakers, on a seamless
      pure-white studio background, $9{:}16$ full-body frame, shot on
      Canon R5 $85$mm.
    \end{minipage}\\[-2pt]
    \includegraphics[width=\linewidth,keepaspectratio]{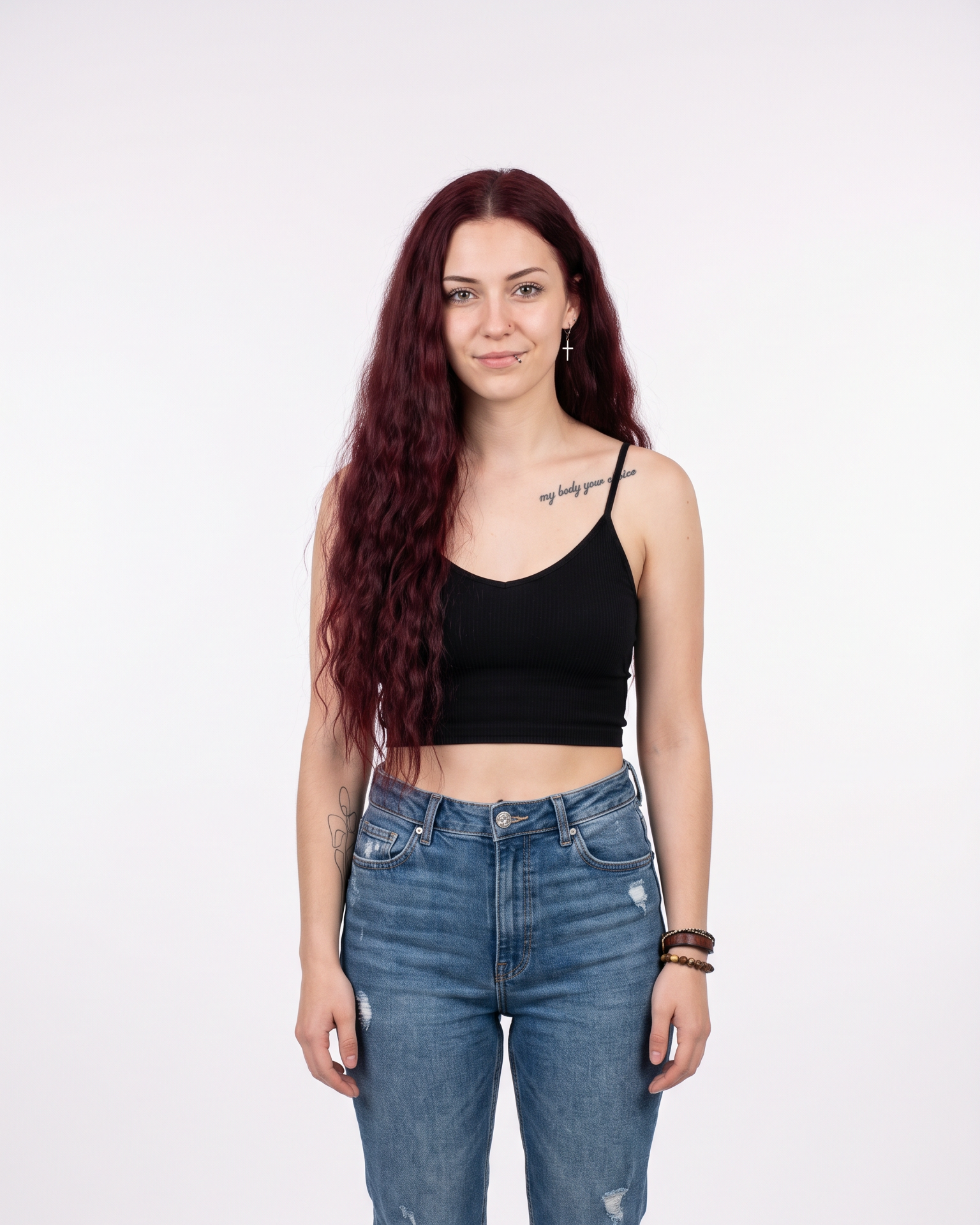}
    \caption*{\footnotesize\textbf{74 questions} \,$\cdot$\, faithfulness $\mathbf{0.768}$
              \quad \scriptsize($\sim 17$ constraints failed)}
  \end{subfigure}\hfill
  \begin{subfigure}{0.49\linewidth}
    \centering
    \begin{minipage}{\linewidth}
      \scriptsize\itshape
      \textbf{\upshape Prompt} (excerpt; \textbf{\upshape category:}
      photorealistic): A hyperrealistic candid smartphone photo of
      Whiterun in central Skyrim, circa 4E~$201$, from the Plains
      District cobblestones looking up at the three-tiered city.
      Foreground: \textbf{\upshape Warmaiden's} smithy with Adrianne
      at the forge, the Banned Mare inn sign, market stalls, a
      Whiterun guard in winged helmet. Mid: \textbf{\upshape Jorrvaskr}
      mead hall like an upturned longship, the Skyforge ring of stone,
      bare-branched Gildergreen. Back: \textbf{\upshape Dragonsreach}
      on the Cloud District summit. Golden-hour light, slight haze, a
      thumb intruding at the frame corner.
    \end{minipage}\\[-2pt]
    \includegraphics[width=\linewidth,keepaspectratio]{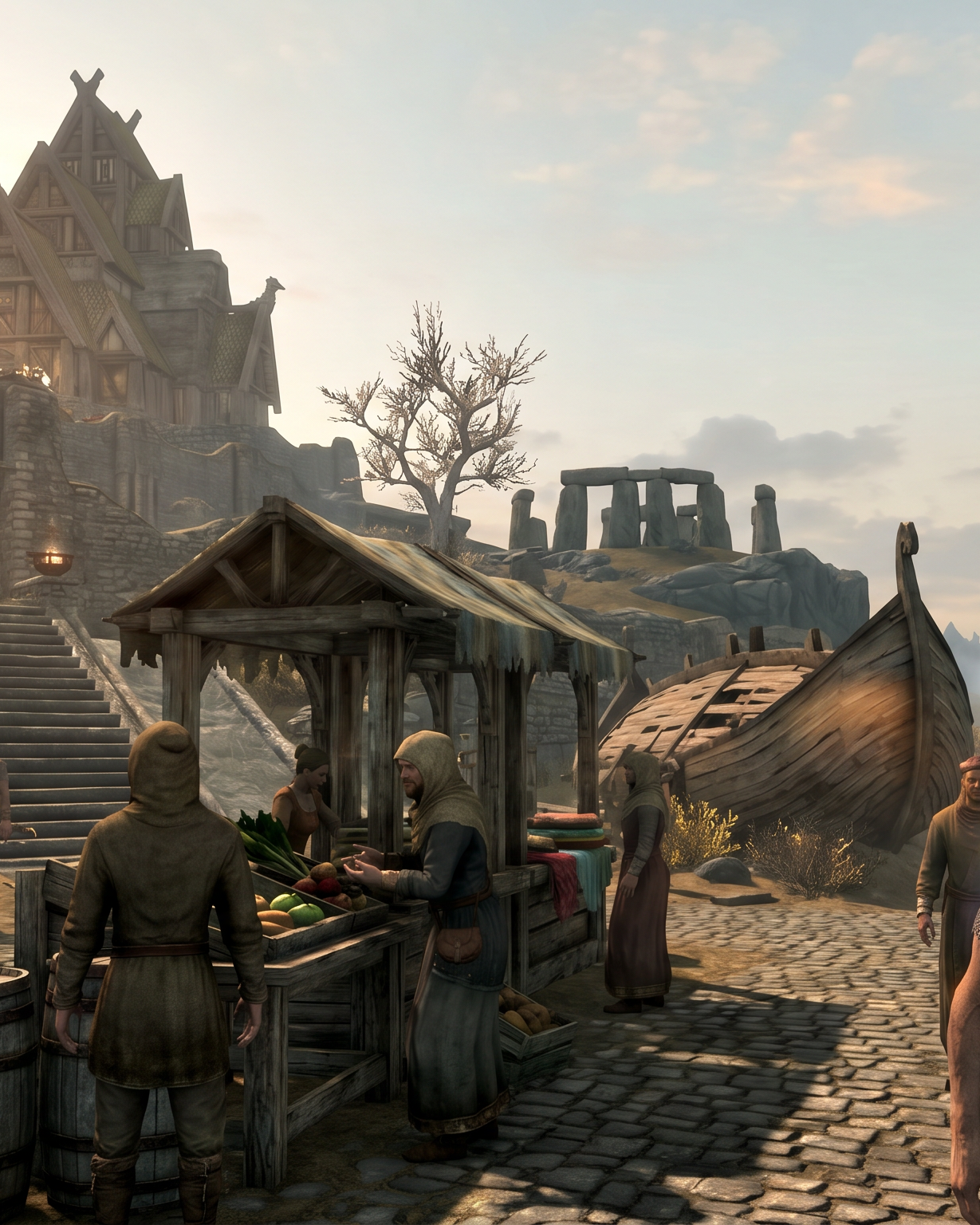}
    \caption*{\footnotesize\textbf{67 questions} \,$\cdot$\, faithfulness $\mathbf{0.857}$
              \quad \scriptsize($\sim 10$ constraints failed)}
  \end{subfigure}
  \caption{\textbf{Two representative Arena-T2I Hard prompts and the
  outputs of \texttt{gemini-3-pro-image-preview-2k}.} Each prompt
  stacks dozens of independently checkable constraints---identity
  details, named landmarks, spatial relationships, lighting, and
  stylistic register. On the portraits prompt the model misses
  $\sim$$17$ of the $74$ constraints (e.g., the ``Polycoria''
  four-pupil detail, the exact tattoo text, the lip ring); on the
  Skyrim cityscape it misses $\sim$$10$ of $67$ (e.g., the Skyforge
  ring of stone, the thumb intrusion in the frame). Images are
  slightly cropped from $1536{\times}2752$ and $2816{\times}1536$
  originals to a common $4{:}5$ portrait aspect for display.}
  \label{fig:arena-hard-examples}
\end{figure}

\paragraph{Qualitative examples.}
Figure~\ref{fig:arena-hard-examples} shows two representative
Arena-T2I Hard prompts and the corresponding outputs from the
strongest closed-source system we evaluate
(\texttt{gemini-3-pro-\allowbreak image-preview-2k}). Even this top system fails
$10$--$17$ of $67$--$74$ decomposed constraints per prompt; the
specific failures (the ``Polycoria'' four-pupil detail in the
portrait, the Skyforge ring of stone and the thumb intrusion at the
corner of the Whiterun cityscape) are the kind of fine-grained
constraints a single binary preference score cannot surface.

\paragraph{Construction statistics.}
Table~\ref{tab:arena-t2i-hard-stats} reports the resulting prompt-level
distribution. Compared to the training set, Arena-T2I Hard prompts are
on average $\sim$$4\times$ longer ($2{,}522$ chars vs.\ $617$) and
carry $\sim$$70\%$ more faithfulness questions ($30.8$ vs.\ $18.2$ per
prompt on average), with dependency-graph depth up to $5$ and as many
as $10$ direct dependencies on a single attribute or relation question.

\begin{table}[t]
  \centering
  \small
  \begin{tabular}{lrrrrrr}
  \toprule
  & \textbf{mean} & \textbf{median} & \textbf{std} & \textbf{p25} & \textbf{p75} & \textbf{max} \\
  \midrule
  \multicolumn{7}{l}{\emph{Prompt length (characters)}} \\
  Training set ($10$k)    & $617$  & $396$  & $884$  & $211$ & $716$ & $17{,}467$ \\
  Arena-T2I Hard ($310$)  & $\mathbf{2{,}522}$ & $\mathbf{1{,}714}$ & $2{,}997$ & $864$  & $2{,}788$ & $18{,}411$ \\
  \midrule
  \multicolumn{7}{l}{\emph{Faithfulness questions per prompt}} \\
  Training set ($10$k)    & $18.2$ & $17$ & $10.8$ & $10$ & $25$ & $70$ \\
  Arena-T2I Hard ($310$)  & $\mathbf{30.8}$ & $\mathbf{31}$ & $12.4$ & $22$ & $40$ & $88$ \\
  \bottomrule
  \end{tabular}
  \caption{Per-prompt structural statistics for Arena-T2I Hard versus the
  $10$k training set, both drawn from the same filtered arena source pool.
  The selection rule favours long prompts with many decomposed
  questions, so Arena-T2I Hard sits roughly $4\times$ above the training
  distribution on prompt length and $\sim$$70\%$ above on question
  count.}
  \label{tab:arena-t2i-hard-stats}
\end{table}

\paragraph{Category coverage.}
Each of the $310$ prompts carries six boolean visual-style category
flags (\emph{commercial\_design}, \emph{3d\_modeling}, \emph{cartoon},
\emph{photorealistic}, \emph{art}, \emph{portraits}) from
\texttt{benchmark\_hard.json}. We additionally derive a seventh
\emph{text} category from question content: a prompt's text flag is on
if at least one of its decomposed yes/no questions explicitly probes
text rendering (questions whose surface form mentions
text/word/letter/font/inscription/sign/caption/logo and similar);
within the primary-category column, a prompt is taken to be
text-primary when more than $30\%$ of its questions are
text-rendering questions, overriding the default visual-style label.
The selection rule (long prompt, many decomposed questions) is
structural and does not target any particular style, but the
resulting set is reasonably balanced across the seven classes:
Table~\ref{tab:arena-t2i-hard-categories} reports both the
primary-category distribution (one label per prompt) and the
multi-label flag distribution (boolean OR across prompts). Two
observations: \emph{(i)}~the seven primary classes range from
$21$ prompts (\emph{text}) to $59$ prompts (\emph{3d\_modeling}) of
$310$, so per-category faithfulness numbers can be reported with
reasonable sample size; and \emph{(ii)}~$287 / 310$ prompts ($92.6\%$)
are multi-labeled across $\geq 2$ flags once \emph{text} is added,
with \emph{cartoon}, \emph{photorealistic}, and \emph{text} each at
$\sim$$50\%$, reflecting the long, scene-rich character of the
selected prompts and the strong overlap between text rendering and
the other visual styles.

\begin{table}[t]
  \centering
  \small
  \begin{tabular}{lrr@{\hspace{1.5em}}lrr}
  \toprule
  \multicolumn{3}{c}{\textbf{Primary category} (one per prompt)}
    & \multicolumn{3}{c}{\textbf{Multi-label flag} (boolean OR)} \\
  \cmidrule(lr){1-3}\cmidrule(lr){4-6}
  Category & Count & \% & Flag & ON & \% \\
  \midrule
  3d\_modeling      & $59$ & $19.0\%$ & cartoon              & $156$ & $50.3\%$ \\
  art               & $56$ & $18.1\%$ & photorealistic       & $152$ & $49.0\%$ \\
  cartoon           & $53$ & $17.1\%$ & text                 & $152$ & $49.0\%$ \\
  photorealistic    & $46$ & $14.8\%$ & 3d\_modeling         & $121$ & $39.0\%$ \\
  portraits         & $45$ & $14.5\%$ & art                  & $100$ & $32.3\%$ \\
  commercial\_design& $30$ &  $9.7\%$ & commercial\_design   &  $87$ & $28.1\%$ \\
  text              & $21$ &  $6.8\%$ & portraits            &  $84$ & $27.1\%$ \\
  \midrule
  \textbf{total}    & $\mathbf{310}$ & $\mathbf{100\%}$
                    & \multicolumn{3}{l}{$287 / 310$ prompts multi-labeled ($92.6\%$)} \\
  \bottomrule
  \end{tabular}
  \caption{Category coverage of Arena-T2I Hard. \textbf{Left:}
  primary-category distribution (the dominant category for each
  prompt; exactly one per prompt; \emph{text} overrides the
  visual-style \texttt{primary\_category} when more than $30\%$ of the
  prompt's questions are text-rendering questions). \textbf{Right:}
  multi-label flag distribution (each flag is independently
  true/false; a prompt may carry multiple). The seven classes range
  from $21$ to $59$ prompts under the primary label, and once
  \emph{text} is added most prompts ($92.6\%$) carry at least two
  flags.}
  \label{tab:arena-t2i-hard-categories}
\end{table}

\paragraph{Per-category faithfulness.}
Table~\ref{tab:arena-t2i-hard-per-category} reports the per-category
faithfulness yes-ratio on Arena-T2I Hard, broken down by the
seven-class \texttt{primary\_category} partition of
Table~\ref{tab:arena-t2i-hard-categories}, for the same $11$
closed-source systems that appear in
Table~\ref{tab:arena-t2i-hard-leaderboard}. Three patterns are worth
noting. \emph{(i)} The overall ranking is largely stable across
categories: the top three (gemini-3-pro, grok-imagine, gpt-image-1.5)
lead in five of the seven categories, and the bottom two
(\texttt{hunyuan-image-3.0}, \texttt{ideogram-v3}) trail in all
seven. \emph{(ii)} \emph{Cartoon} is the hardest category for nearly
every model (e.g., gemini-3-pro drops from $0.855$ overall to $0.816$
on cartoon; gpt-image-1.5 from $0.796$ to $0.767$), suggesting
cartoon-style prompts pose extra compositional challenges (stylised
proportions, multiple characters, tightly stacked attributes).
\emph{(iii)} \emph{Text} is the most discriminating axis: the spread
across the eleven systems is $0.471$--$0.906$ ($43$~pp), wider than
on any other primary category, with \texttt{hunyuan-image-3.0} at
$0.471$ ($\sim$$14$~pp below its overall) at the bottom and
\texttt{gemini-3-pro} at $0.906$ ($\sim$$5$~pp above its overall) at
the top. The leaderboard also shuffles within categories:
\texttt{recraft-v4} is the strongest on \emph{portraits} ($0.881$,
ahead of gemini-3-pro at $0.835$), and \texttt{grok-imagine} is the
strongest on \emph{photorealistic} ($0.875$, ahead of gemini-3-pro at
$0.847$). No single closed-source system dominates every category.

\begin{table}[t]
  \centering
  \small
  \setlength{\tabcolsep}{3pt}
  \begin{tabular}{lrrrrrrr|r}
  \toprule
  \textbf{Model}
    & \textbf{3D}
    & \textbf{Art}
    & \textbf{Cart.}
    & \textbf{Com.}
    & \textbf{Photo.}
    & \textbf{Port.}
    & \textbf{Text}
    & \textbf{Overall} \\
  \midrule
  \texttt{gemini-3-pro-image-preview-2k}        & $\mathbf{0.862}$ & $\mathbf{0.896}$ & $\mathbf{0.816}$ & $0.842$        & $0.847$        & $0.835$        & $\mathbf{0.906}$ & $\mathbf{0.855}$ \\
  \texttt{grok-imagine-image-20260306}          & $0.789$        & $0.889$        & $0.803$        & $\mathbf{0.855}$ & $\mathbf{0.875}$ & $0.875$        & $0.881$        & $0.849$ \\
  \texttt{gpt-image-1.5-high-fidelity}          & $0.820$        & $0.811$        & $0.767$        & $0.753$        & $0.831$        & $0.797$        & $0.744$        & $0.796$ \\
  \texttt{recraft-v4}                           & $0.720$        & $0.766$        & $0.737$        & $0.812$        & $0.820$        & $\mathbf{0.881}$ & $0.817$        & $0.787$ \\
  \texttt{wan2.6-t2i-v2}                        & $0.733$        & $0.779$        & $0.737$        & $0.715$        & $0.790$        & $0.839$        & $0.800$        & $0.768$ \\
  \texttt{gemini-2.5-flash-image} (nano-banana) & $0.761$        & $0.781$        & $0.741$        & $0.754$        & $0.782$        & $0.790$        & $0.750$        & $0.767$ \\
  \texttt{gpt-image-1}                          & $0.705$        & $0.778$        & $0.686$        & $0.658$        & $0.739$        & $0.742$        & $0.726$        & $0.722$ \\
  \texttt{imagen-4.0-ultra-generate-001}        & $0.661$        & $0.693$        & $0.666$        & $0.577$        & $0.691$        & $0.773$        & $0.659$        & $0.680$ \\
  \texttt{imagen-4.0-generate-001}              & $0.647$        & $0.683$        & $0.649$        & $0.661$        & $0.654$        & $0.687$        & $0.602$        & $0.659$ \\
  \texttt{hunyuan-image-3.0-fal}                & $0.618$        & $0.625$        & $0.566$        & $0.510$        & $0.666$        & $0.700$        & $0.471$        & $0.609$ \\
  \texttt{ideogram-v3-quality}                  & $0.526$        & $0.515$        & $0.508$        & $0.450$        & $0.520$        & $0.578$        & $0.570$        & $0.523$ \\
  \bottomrule
  \end{tabular}
  \caption{Per-category faithfulness yes-ratio on Arena-T2I Hard for
  the $11$ closed-source systems of Table~\ref{tab:arena-t2i-hard-leaderboard},
  broken down by the seven-class primary partition of
  Table~\ref{tab:arena-t2i-hard-categories}. Columns: \emph{3D} =
  \texttt{3d\_modeling}, \emph{Art} = \texttt{art}, \emph{Cart.} =
  \texttt{cartoon}, \emph{Com.} = \texttt{commercial\_design},
  \emph{Photo.} = \texttt{photorealistic}, \emph{Port.} =
  \texttt{portraits}, \emph{Text} = prompts where more than $30\%$ of
  the questions probe text rendering. Per-category sample sizes are
  $59 / 56 / 53 / 30 / 46 / 45 / 21$ prompts respectively (cf.\
  Table~\ref{tab:arena-t2i-hard-categories}). Bold marks the
  per-column best.}
  \label{tab:arena-t2i-hard-per-category}
\end{table}

\paragraph{Closed-source leaderboard.}
The full closed-source leaderboard is reported in
Table~\ref{tab:arena-t2i-hard-leaderboard} of the introduction. Two
points worth pulling out for the appendix reader. \emph{First}, the
ranking on Arena-T2I Hard correlates only loosely with the public arena
leaderboard \emph{in both directions}: several public top-$15$ systems
drop to the bottom half of our ranking
(\texttt{hunyuan-image-3.0} \#$15{\to}10$th, \texttt{imagen-4.0-ultra}
\#$17{\to}8$th, \texttt{imagen-4.0} \#$22{\to}9$th), while
\texttt{recraft-v4} climbs from \#$29$ publicly to $4$th on
faithfulness and \texttt{ideogram-v3-quality} drops from \#$42$ to
last. \emph{Second}, our strongest open-source fine-tune (FLUX.1-dev
with the Faith\,$+$\,Pick recipe at $0.529$) is comparable to the
bottom of the closed-source ladder, which is why we report the main
MMRB2 head-to-head results
(Section~\ref{sec:results:main}) on the easier 1k test set rather than
on Arena-T2I Hard.

\paragraph{Decomposer robustness.}
The Arena-T2I Hard leaderboard depends on the choice of offline
decomposer $\mathcal{G}$. To check that the ranking is not an artefact
of using Gemini-3-Pro, we re-decompose all $310$ prompts with a second
decomposer---GPT-5.4---and re-score the same $11$ closed-source systems
under the same gemini-3-flash judge. The two decomposers produce
question graphs of substantially different size: Gemini-3-Pro emits
$\sim$$10$ questions per prompt on average, GPT-5.4 emits
$\sim$$43$ ($\sim$$4{\times}$ more), so we expect every system's
yes-ratio to drop uniformly under GPT-5.4 (every system has more
questions to satisfy). The headline result is that the
\emph{ranking is essentially preserved}:
\textbf{Pearson} $0.991$ and \textbf{Spearman} $0.982$
on the per-system mean yes-rate vector between the two decomposers
(Table~\ref{tab:decomposer-robustness}). The absolute scores all drop
by $0.05$--$0.09$ as expected, but no system flips its rank by more
than one position. We therefore attribute the leaderboard signal to
the prompts themselves and not to the choice of decomposer. (At the
finer per-prompt level the agreement is more moderate---per-prompt
$11$-vector correlation has mean Pearson $0.67$ / Spearman
$0.68$---which is consistent with two decomposers asking different
specific questions about the same prompt while the
many-questions-per-prompt aggregation washes out the noise.)

\begin{table}[t]
  \centering
  \footnotesize
  \setlength{\tabcolsep}{4pt}
  \begin{tabular}{rlrrr}
  \toprule
  \textbf{\#} & \textbf{Model}
              & \textbf{Gemini-3-Pro} $\mathcal{G}$
              & \textbf{GPT-5.4} $\mathcal{G}$
              & $\Delta$ \\
  \midrule
  $1$ & \texttt{gemini-3-pro-image-preview-2k}        & $0.866$ & $0.802$ & $-0.064$ \\
  $2$ & \texttt{grok-imagine-image-20260306}          & $0.850$ & $0.803$ & $-0.047$ \\
  $3$ & \texttt{gpt-image-1.5-high-fidelity}          & $0.801$ & $0.729$ & $-0.072$ \\
  $4$ & \texttt{recraft-v4}                           & $0.794$ & $0.723$ & $-0.071$ \\
  $5$ & \texttt{wan2.6-t2i-v2}                        & $0.770$ & $0.683$ & $-0.087$ \\
  $6$ & \texttt{gemini-2.5-flash-image} (nano-banana) & $0.762$ & $0.714$ & $-0.048$ \\
  $7$ & \texttt{gpt-image-1}                          & $0.726$ & $0.648$ & $-0.078$ \\
  $8$ & \texttt{imagen-4.0-ultra-generate-001}        & $0.681$ & $0.633$ & $-0.048$ \\
  $9$ & \texttt{imagen-4.0-generate-001}              & $0.670$ & $0.587$ & $-0.083$ \\
 $10$ & \texttt{hunyuan-image-3.0-fal}                & $0.620$ & $0.557$ & $-0.063$ \\
 $11$ & \texttt{ideogram-v3-quality}                  & $0.518$ & $0.454$ & $-0.064$ \\
  \midrule
  \multicolumn{2}{l}{\emph{Avg q/prompt}}
              & $\sim 10$ & $\sim 43$ & --- \\
  \multicolumn{2}{l}{\emph{Cross-decomposer rank corr.}}
              & \multicolumn{2}{c}{\textbf{Pearson $\mathbf{0.991}$}, \textbf{Spearman $\mathbf{0.982}$}} & \\
  \bottomrule
  \end{tabular}
  \caption{Decomposer-robustness check on Arena-T2I Hard. Each cell is
  the per-system mean yes-rate, computed over the
  $243$ prompts for which both decomposers' question graphs scored
  successfully. Numbers are the headline mean over those $243$
  prompts; they are slightly different from
  Table~\ref{tab:arena-t2i-hard-leaderboard} (which uses all
  $310$ prompts under Gemini-3-Pro only) because of this
  intersection. GPT-5.4 produces $\sim$$4{\times}$ more questions
  per prompt than Gemini-3-Pro, which uniformly lowers every
  system's yes-rate by $0.05$--$0.09$, but the system ranking is
  essentially preserved (no rank flip greater than one position).}
  \label{tab:decomposer-robustness}
\end{table}

\paragraph{Open-source comparison.}
Table~\ref{tab:arena-t2i-hard-trained} reports the same faithfulness
yes-ratio on every open-source backbone and BT/SFT fine-tune we ran in
this paper, on both Arena-T2I Hard ($310$ prompts) and DPG-Bench
($1{,}065$ prompts). Our combined Faith\,$+$\,Pick recipe is at the
top of the ranking on \emph{both} benchmarks and \emph{both} backbones:
SD3.5-M reaches $0.405$ on Arena-T2I Hard ($+7.8$~pp over base) and
$0.915$ on DPG-Bench ($+2.1$~pp), and FLUX.1-dev reaches $0.529$ on
Arena-T2I Hard ($+8.3$~pp) and $0.935$ on DPG-Bench ($+3.3$~pp). Two
side-observations agree with the MMRB2 main-results matrices in
Section~\ref{sec:results:main}: on SD3.5-M, HPSv3 ($0.281$) and the
4-reward ensemble ($0.322$) both score \emph{below} the base backbone
($0.328$) on Arena-T2I Hard---and the same below-base ordering
reproduces on DPG-Bench (HPSv3 $0.875$, ensemble $0.887$, base
$0.894$)---echoing the $32.9\%$ and $38.2\%$ MMRB2 win rates against
base reported in §\ref{sec:results:main}; and DreamSync, a non-RL
baseline, sits well below Faith\,$+$\,Pick on both backbones. On
DPG-Bench the BT-only spread is much narrower than on Arena-T2I Hard
(SD3.5-M: $\sim$$2$~pp on DPG vs.\ $\sim$$7$~pp on Arena-T2I Hard;
FLUX.1-dev: $\sim$$1$~pp on DPG vs.\ $\sim$$10$~pp on Arena-T2I Hard)
because DPG is closer to saturation, consistent with the closed-source
pattern in Table~\ref{tab:arena-t2i-hard-leaderboard}.

\begin{table}[t]
  \centering
  \small
  \begin{tabular}{rllrr}
  \toprule
  \textbf{\#} & \textbf{Backbone} & \textbf{Method}
              & \textbf{Arena-T2I Hard $\uparrow$}
              & \textbf{DPG-Bench $\uparrow$} \\
  \midrule
  $1$ & SD3.5-M & \emph{Faith\,$+$\,Pick} (ours)        & $\mathbf{0.405}$ & $\mathbf{0.915}$ \\
  $2$ & SD3.5-M & PickScore                             & $0.353$          & $0.894$ \\
  $3$ & SD3.5-M & DreamSync (iter 1)                    & $0.351$          & $0.891$ \\
  $4$ & SD3.5-M & ImageReward                           & $0.331$          & $0.894$ \\
   -- & SD3.5-M & \emph{Base SD3.5-M (no fine-tune)}    & $0.328$          & $0.894$ \\
  $5$ & SD3.5-M & 4-reward ensemble                     & $0.322$          & $0.887$ \\
  $6$ & SD3.5-M & HPSv3                                 & $0.281$          & $0.875$ \\
  \midrule
  $1$ & FLUX.1-dev & \emph{Faith\,$+$\,Pick} (ours)        & $\mathbf{0.529}$ & $\mathbf{0.935}$ \\
  $2$ & FLUX.1-dev & DreamSync (iter 2)                    & $0.492$          & $0.918$ \\
  $3$ & FLUX.1-dev & HPSv3                                 & $0.468$          & $0.909$ \\
  $4$ & FLUX.1-dev & PickScore                             & $0.463$          & $0.919$ \\
  $5$ & FLUX.1-dev & 4-reward ensemble                     & $0.461$          & $0.906$ \\
   -- & FLUX.1-dev & \emph{Base FLUX.1-dev (no fine-tune)} & $0.446$          & $0.902$ \\
  $6$ & FLUX.1-dev & ImageReward                           & $0.425$          & $0.910$ \\
  \bottomrule
  \end{tabular}
  \caption{Open-source models on Arena-T2I Hard ($310$ prompts) and
  on DPG-Bench ($1{,}065$ prompts), grouped by backbone and ranked by
  Arena-T2I Hard faithfulness yes-ratio (same gemini-3-flash judge
  on both benchmarks, matching
  Table~\ref{tab:arena-t2i-hard-leaderboard}). Italic rows are the
  untrained backbones, included as reference points; the $\#$ column
  ranks fine-tuned variants only. On both backbones,
  Faith\,$+$\,Pick is the top recipe on DPG-Bench as well as on
  Arena-T2I Hard (SD3.5-M: $0.915$; FLUX.1-dev: $0.935$, $+3.3$~pp
  over base). The BT-only spread on DPG is much narrower than on
  Arena-T2I Hard ($\sim$$2$~pp on SD3, $\sim$$1$~pp on FLUX, vs.\
  $\sim$$7$~pp and $\sim$$10$~pp respectively) because DPG is closer
  to saturation.}
  \label{tab:arena-t2i-hard-trained}
\end{table}

We do not use Arena-T2I Hard for the main MMRB2 head-to-head results in
Section~\ref{sec:results:main}; we release it as a stress benchmark for
assessing the faithfulness ceiling of stronger T2I systems.

\subsection{Provenance and licensing}
\label{app:bench:provenance}

The prompt source license permits redistribution for research purposes; we will
release the prompt JSON and the decomposed questions under the same license.

\section{Implementation Details}
\label{app:impl}

This appendix expands the implementation details that did not fit in
Section~\ref{sec:reward}.

\subsection{Decomposition prompts}
\label{app:impl:decompose}

We decompose each training prompt offline with two text-only Gemini-3-Pro calls:
the first produces faithfulness questions, the second adds an aesthetics layer
that references the faithfulness ids. Results are cached on disk and never
recomputed at training time.

\paragraph{Faithfulness decomposition.}
The first pass enforces (i)~one-attribute-per-question, (ii)~the dependency
rules described in Section~\ref{sec:reward:design}, and (iii)~a strict JSON
output schema. The system prompt is reproduced verbatim below.

\begin{Verbatim}
You are an expert at analyzing text-to-image generation prompts. Your job
is to decompose a complex image generation prompt into a list of simple
yes/no verification questions WITH dependency information. These questions
will later be used to check whether a generated image is faithful to the
original prompt.

Guidelines:
- Each question should check exactly ONE visual attribute (object
  existence, color, spatial relationship, count, action, style, text
  content, etc.).
- Questions must be answerable by looking at the image alone (given the
  original prompt for context).
- Use clear, unambiguous language.
- Cover ALL important details mentioned in the prompt. Do not skip
  anything.
- Order questions from most important (core subject / objects) to least
  important (minor stylistic details).
- Do NOT ask about things not mentioned or implied by the prompt.

Dependency rules:
- Start with existence questions for each key object (e.g. "Is there a
  robot?").
- Attribute questions (color, style, pose, action) about an object MUST
  depend on that object's existence question.
- Relationship questions between two objects depend on BOTH objects'
  existence.
- A question can depend on multiple parent questions (list all parent ids).
- Root questions (no dependency) have "depends_on": [].

Output format:
Return ONLY a JSON array of objects. Each object has:
  - "id": integer starting from 0
  - "question": the yes/no question string
  - "depends_on": list of integer ids that this question depends on
    (empty for root)

Example for "A red cat sitting on a blue chair":
[
  {"id": 0, "question": "Is there a cat in the image?",
   "depends_on": []},
  {"id": 1, "question": "Is the cat red?", "depends_on": [0]},
  {"id": 2, "question": "Is there a chair in the image?",
   "depends_on": []},
  {"id": 3, "question": "Is the chair blue?", "depends_on": [2]},
  {"id": 4, "question": "Is the cat sitting on the chair?",
   "depends_on": [0, 2]}
]

No explanation, no markdown fences. ONLY the JSON array.
\end{Verbatim}

\paragraph{Aesthetics decomposition.}
The second pass receives the faithfulness questions as context and emits a
parallel set of aesthetic-quality questions whose ids continue from where the
faithfulness ids left off. Each aesthetic question is framed so that ``yes''
$=$ aesthetically good. The system prompt is reproduced verbatim below.

\begin{Verbatim}
You are an expert at evaluating the aesthetic quality of AI-generated
images. You will be given a text-to-image prompt AND a list of
faithfulness check questions that identify the key visual components in
the image. Your job is to generate a list of yes/no aesthetic quality
questions that evaluate how WELL each component is rendered, and how
harmoniously they work together.

Your questions should cover these aesthetic dimensions for each relevant
component:
- Rendering quality: Is the component rendered with fine detail, proper
  proportions, and realistic/stylistically consistent appearance?
- Color harmony: Do the colors of this component look natural and
  harmonious with the rest of the image?
- Lighting & shading: Is the lighting on this component consistent and
  visually appealing?
- Composition: Is this component well-placed within the overall image
  composition?
- Overall coherence: Do all components work together to create a visually
  pleasing and coherent scene?

Guidelines:
- Each question should check exactly ONE aesthetic aspect of ONE
  component or the overall image.
- Frame ALL questions so that "yes" = aesthetically good, "no" =
  aesthetically poor.
- Questions must be answerable by looking at the image alone.
- Use the faithfulness questions to understand what components exist in
  the image. Reference specific components from the prompt (e.g., "the
  cat", "the background").
- Include dependencies: aesthetic questions about a component depend on
  that component's existence question from the faithfulness list.
- Always include global questions (composition, overall harmony) as root
  questions with no dependencies.
- Do NOT repeat the faithfulness questions. Focus ONLY on aesthetic
  quality.

Input format:
You will receive:
1. The original prompt.
2. A JSON list of faithfulness questions (with id, question, depends_on).

Output format:
Return ONLY a JSON array of objects. Each object has:
  - "id": integer (continue numbering from the last faithfulness question
    id + 1)
  - "question": the yes/no aesthetic question string ("yes" = good)
  - "depends_on": list of integer ids from the faithfulness questions
    that this question depends on (use the faithfulness question ids for
    component existence)

Example - given faithfulness questions for "A red cat sitting on a blue
chair":
[
  {"id": 5, "question": "Is the overall image composition well-balanced
   and visually appealing?", "depends_on": []},
  {"id": 6, "question": "Is the cat rendered with fine detail, proper
   anatomy, and realistic fur texture?", "depends_on": [0]},
  {"id": 7, "question": "Does the red color of the cat look natural and
   visually harmonious with the scene?", "depends_on": [0]},
  {"id": 8, "question": "Is the chair rendered with clean lines, proper
   perspective, and convincing material texture?", "depends_on": [2]},
  {"id": 9, "question": "Does the blue color of the chair complement the
   overall color palette of the image?", "depends_on": [2]},
  {"id": 10, "question": "Is the lighting across the scene consistent and
   does it create appealing highlights and shadows?",
   "depends_on": []},
  {"id": 11, "question": "Do the cat and chair look naturally integrated
   together without awkward boundaries or scale issues?",
   "depends_on": [0, 2]}
]

No explanation, no markdown fences. ONLY the JSON array.
\end{Verbatim}

\subsection{VLM judge prompts}
\label{app:impl:vlm}

Two system prompts are used at training time, one per query mode. In both
modes the judge sees the original T2I prompt, the generated image, and the
question(s) at hand. \textsc{irrelevant} answers are mapped to $0$ in the
default reward; we extract the response with a case-insensitive regex match
against the three labels and treat malformed responses as \textsc{irrelevant}.

\paragraph{Per-question (\emph{individual}) mode.}
\begin{Verbatim}
You are an impartial image quality judge. You will be given:
1. The original text-to-image prompt.
2. A specific yes/no verification question about the image.
3. The generated image.

Your task: look at the image and answer the question.

Rules:
- Answer with exactly one word: "yes", "no", or "irrelevant".
  - "yes"        = the image clearly satisfies the question.
  - "no"         = the image clearly does NOT satisfy the question.
  - "irrelevant" = the question does not apply to this image or cannot
    be determined from the image.
- Do NOT explain. Output ONLY one of the three words.
\end{Verbatim}

\paragraph{Oneshot mode.}
\begin{Verbatim}
You are an impartial image quality judge. You will be given:
1. The original text-to-image prompt.
2. A list of yes/no verification questions about the image, each with an
   integer id.
3. The generated image.

Your task: look at the image and answer ALL questions in order.

Rules:
- For each question, answer "yes", "no", or "irrelevant".
  - "yes"        = the image clearly satisfies the question.
  - "no"         = the image clearly does NOT satisfy the question.
  - "irrelevant" = the question does not apply or cannot be determined.
- Output ONLY a JSON array of objects, each with "id" (int) and "answer"
  (string).
- Do NOT explain. No markdown fences. ONLY the JSON array.

Example output:
[{"id": 0, "answer": "yes"}, {"id": 1, "answer": "no"},
 {"id": 2, "answer": "irrelevant"}]
\end{Verbatim}

\subsection{Judge benchmark evaluation}
\label{app:impl:judge-eval}

\begin{figure}[t]
  \centering
  \includegraphics[width=\linewidth]{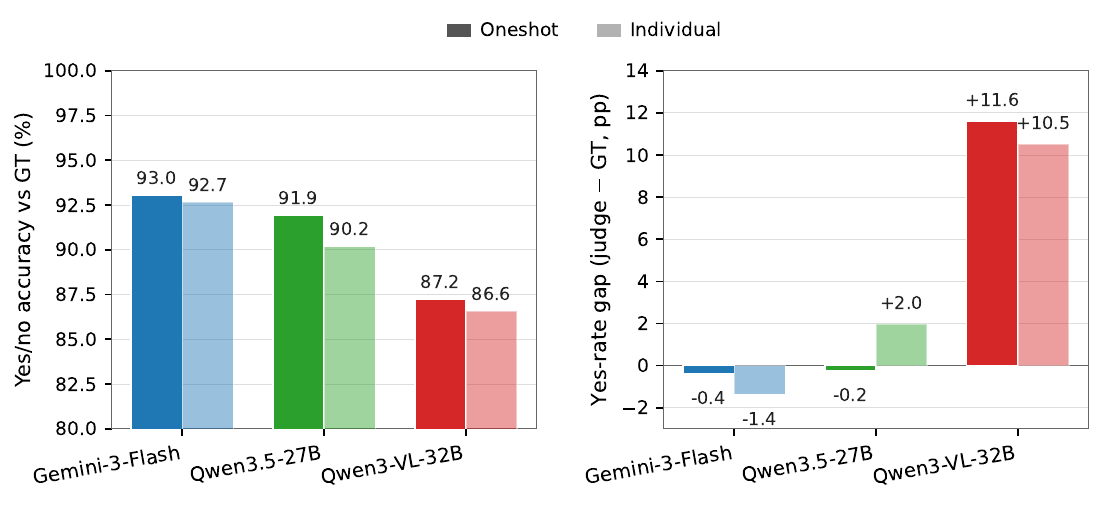}
  \caption{VLM-judge faithfulness evaluation on the $100$-prompt
  human-labelled benchmark. Each judge family is shown under both query
  modes (solid bar $=$ oneshot, faded bar $=$ individual).
  \textbf{Left}: yes/no accuracy against GT. \textbf{Right}: yes-rate gap
  (judge $-$ GT, in percentage points). The weaker Qwen3-VL-32B both
  loses ${\sim}5$~pp of accuracy and over-predicts \textsc{yes} by
  ${\sim}6{\times}$ more than the other two judges; oneshot vs.\
  individual is within $1.7$~pp on accuracy for every family.}
  \label{fig:judge-eval}
\end{figure}

This appendix expands Section~\ref{sec:reward:impl}'s judge selection.
The benchmark consists of $100$ prompts sampled from real user data, one
SD3-Medium base-model image per prompt, decomposed into $1{,}810$ yes/no
questions in total ($\sim$18 per prompt) and labelled by hand. The final
label distribution is $1{,}268$ yes / $505$ no / $37$ irrelevant.
Figure~\ref{fig:judge-eval} reports two metrics for each judge $\times$
query-mode configuration: yes/no accuracy against the human ground truth
(left), and the gap between the judge's yes-rate and the GT yes-rate
(right). Three observations: (i)~the per-judge accuracy ranking is
Gemini-3-Flash $>$ Qwen3.5-27B $>$ Qwen3-VL-32B; (ii)~oneshot and
individual produce near-identical accuracy within each family (largest
gap $1.7$~pp on Qwen3.5-27B); and (iii)~stronger judges are
well-calibrated to the GT yes-rate ($\pm 2$~pp) while the weaker
Qwen3-VL-32B leans toward \textsc{yes} by ${\sim}11$~pp---a
weaker-models-lean-yes pattern that is also visible in the higher
sensitivity ($0.99$) but much lower specificity ($0.49$) of Qwen3-VL-32B
relative to the other two judges (specificity $0.81$--$0.85$).

\begin{figure}[t]
  \centering
  \includegraphics[width=\linewidth,height=0.78\textheight,keepaspectratio]{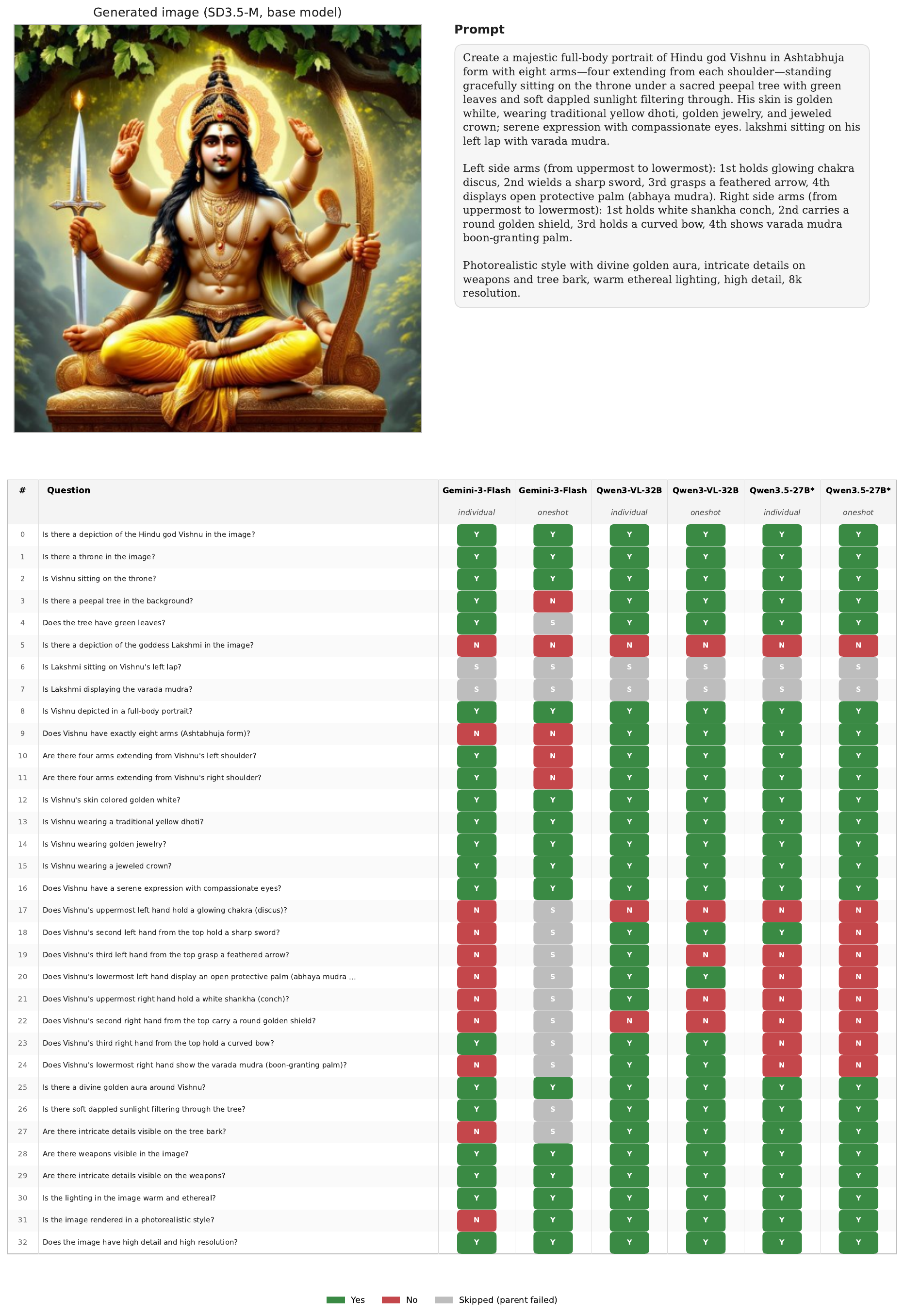}
  \caption{Per-question yes/no/skip answers from six judge configurations
  on a single benchmark prompt (Arena prompt~\#48, SD3.5-M base-model
  image, $33$ decomposed faithfulness questions). Green $=$ yes,
  red $=$ no, grey $=$ skipped because a parent question failed. Six
  columns are three judge families (Gemini-3-Flash, Qwen3-VL-32B,
  Qwen3.5-27B) crossed with two query modes (individual, oneshot).
  Qwen3-VL-32B's higher yes-rate is visible as extra green cells on the
  fine-grained iconographic rows.}
  \label{fig:prompt48-judges}
\end{figure}

To make the weaker-judges-lean-yes pattern concrete at the per-question
level, Figure~\ref{fig:prompt48-judges} renders all six judge
configurations on a single benchmark prompt (Arena prompt~\#48,
$33$ decomposed questions). Each column is one (judge, mode)
configuration; each row is one question; cell colour shows the judge's
answer (green $=$ yes, red $=$ no, grey $=$ skipped because a parent
question failed). Within a column, oneshot and individual answers are
nearly identical---consistent with the per-question agreement reported
in Section~\ref{sec:reward:impl}. Across columns, all three judges agree
on the unambiguous existence and attribute questions, but on
fine-grained iconographic items---which arm holds the sword, the conch,
the bow---Qwen3-VL-32B's higher yes-rate appears as extra green cells
where the human GT and the stronger judges correctly mark \textsc{no}.

\subsection{vLLM serving and endpoint pool}
\label{app:impl:pool}

\paragraph{Serving.}
The vision--language judge $\mathcal{V}$ (Qwen3.5-27B-Instruct) is served as
multiple independent vLLM instances. We spin up one instance per GPU on each
host: a typical configuration is 5 hosts $\times$ 8 GPUs $=$ 40 endpoints, each
exposing the OpenAI-compatible chat-completions API on a distinct port. Each
endpoint serves the same model checkpoint and uses default vLLM tensor-parallel
settings (TP $=$ 1) so a single H200 holds the full 27B model in
\texttt{bf16}. Image inputs are passed as base-64 JPEG data URLs, encoded once
per generation and reused across all questions for that image.

\paragraph{Endpoint registry.}
A JSON registry on disk (\texttt{vllm\_server/endpoints.json}) maps named keys
(\texttt{qwen3.5}, \texttt{aren\_3vl}, $\dots$) to lists of base URLs. The
scorer re-reads the registry on every batch via \texttt{mtime} detection, so
endpoints can be added or drained on a running training job without restarting
it; if the file is missing or malformed, the previously cached list is reused
and a warning is logged.

\paragraph{Load balancing.}
Within a batch, the scorer picks an endpoint by a \emph{least-connections} rule
(always send the next request to the endpoint with the fewest pending
requests), which prevents slow endpoints from causing head-of-line blocking. A
per-endpoint semaphore (default \texttt{max\_concurrent\_per\_endpoint}$=$1)
keeps any single vLLM instance from queueing more than one request at a time;
the global thread-pool size auto-adapts to
$(\#\text{endpoints}) \times \texttt{max\_concurrent\_per\_endpoint}$. Failed
calls (timeout, dead endpoint, malformed response) trigger up to five retries
with exponential backoff and a fresh endpoint pick at each retry; if every
retry fails, the question's score is set to a sentinel and excluded from
aggregation for that image.

\subsection{MMRB2 evaluation prompt}
\label{app:eval}

For the pairwise win-rate evaluation in
Section~\ref{sec:experiments}, we feed the Gemini-3-flash judge the
text-to-image rubric below verbatim from the MMRB2 release, together with the
original T2I prompt and two candidate images (labeled \texttt{Response A} and
\texttt{Response B}). The judge returns structured JSON containing per-criterion
reasoning, an integer score in $\{1,\dots,6\}$, a \texttt{better\_response}
verdict, and a confidence estimate. We use only the \texttt{better\_response}
field to compute the net win rate; each pair is judged twice with the order
swapped to remove position bias.

\begin{Verbatim}
You are an expert in multimodal quality analysis and generative AI
evaluation. Your role is to act as an objective judge for comparing
two AI-generated responses to the same prompt. You will evaluate
which response is better based on a comprehensive rubric.

**Important Guidelines:**
- Be completely impartial and avoid any position biases
- Ensure that the order in which the responses were presented does
  not influence your decision
- Do not allow the length of the responses to influence your
  evaluation
- Do not favor certain model names or types
- Be as objective as possible in your assessment
- Consider factors such as helpfulness, relevance, accuracy, depth,
  creativity, and level of detail

**Understanding the Content Structure:**
- **[ORIGINAL PROMPT TO MODEL:]**: instruction given to both models
- **[INPUT IMAGE FROM PROMPT:]**: source image provided (if any)
- **[RESPONSE A:]**: first model's generated response
- **[RESPONSE B:]**: second model's generated response

Your evaluation must be based on a fine-grained rubric covering the
criteria below. For each criterion, provide detailed step-by-step
reasoning comparing both responses, on a 1-6 scoring scale.

**Evaluation Criteria:**
1. **faithfulness_to_prompt:** Which response better adheres to the
   composition, objects, attributes, and spatial relationships
   described in the text prompt?
2. **text_rendering:** If either response contains rendered text,
   which has better text quality (spelling, legibility,
   integration)? Otherwise: "Not Applicable."
3. **input_faithfulness:** If an input image is provided, which
   response better respects and incorporates the key elements and
   style of the source? Otherwise: "Not Applicable."
4. **image_consistency:** For multi-image responses, which has
   better visual consistency (character appearance, scene details)?
   Otherwise: "Not Applicable."
5. **text_image_alignment:** Which response has better alignment
   between text descriptions and visual content?
6. **text_quality:** If text was generated, which response has
   better linguistic quality (correctness, coherence, grammar,
   tone)?
7. **overall_quality:** Which response has better general technical
   and aesthetic quality, realism, coherence, and fewer visual
   artifacts or distortions?

**Scoring Rubric:**
- 6: Response A significantly better across most criteria
- 5: Response A marginally better across several criteria
- 4: Unsure / A negligibly better
- 3: Unsure / B negligibly better
- 2: Response B marginally better
- 1: Response B significantly better

**Confidence Assessment** (0.0 - 1.0). Be conservative: default to
0.3-0.5 for most comparisons; reserve 0.6-0.7 for clearly
differentiated cases; use 0.8+ ONLY when one response is
dramatically better across ALL criteria with zero ambiguity (less
than 10% of cases).

**Output format** (single JSON object):
{
    "reasoning": {
        "faithfulness_to_prompt":  "...",
        "text_rendering":          "...",
        "input_faithfulness":      "...",
        "image_consistency":       "...",
        "text_image_alignment":    "...",
        "text_quality":            "...",
        "overall_quality":         "...",
        "comparison_summary":      "..."
    },
    "score":                <int 1-6>,
    "better_response":      "A" | "B",
    "confidence":           <float 0.0-1.0>,
    "confidence_rationale": "..."
}
\end{Verbatim}

\section{Training Setup}
\label{app:training_setup}

\subsection{Hyperparameters}
\label{app:hparams}

Table~\ref{tab:hparams} reports the full training recipe used for our SD3 and FLUX
runs. All RL runs are LoRA fine-tunes of the pretrained backbone; the base weights
remain frozen. Hyperparameters not listed below are inherited from the base config in
the released codebase (\texttt{config/base.py} and \texttt{config/grpo\_faithfulness.py}).

\begin{table}[t]
  \centering
  \caption{Training hyperparameters. \emph{$K$} is the number of rollouts per prompt
  used to compute the group-relative advantage. \emph{Effective prompt batch} is the
  number of unique prompts processed per outer iteration. \emph{KL $\beta$} is the
  coefficient on the KL-to-reference penalty.}
  \label{tab:hparams}
  \small
  \setlength{\tabcolsep}{4pt}
  \begin{tabular}{l c c}
    \toprule
                                & \bfseries SD3.5-M           & \bfseries FLUX.1-dev      \\
    \midrule
    Backbone                    & SD3.5-Medium                 & FLUX.1-dev               \\
    Mixed precision             & fp16                         & bf16                     \\
    Resolution                  & $512 \times 512$             & $512 \times 512$         \\
    LoRA target / rank / $\alpha$ & attention Q/K/V/out, $r{=}32$, $\alpha{=}64$ & attention Q/K/V/out, $r{=}32$, $\alpha{=}64$ \\
    Optimizer                   & AdamW                        & AdamW                    \\
    Learning rate               & $3{\times}10^{-4}$           & $3{\times}10^{-4}$       \\
    Train timesteps $T_{\text{train}}$ & 10                    & 6                        \\
    Eval timesteps $T_{\text{eval}}$   & 40                    & 28                       \\
    CFG scale                   & 4.5                          & 3.5                      \\
    Rollouts per prompt $K$     & 24                           & 24                       \\
    Effective prompt batch      & 48                           & 48                       \\
    Train batch size (per GPU)  & 9                            & 3                        \\
    Test batch size             & 16                           & 16                       \\
    Gradient accumulation steps & $N/2$ (auto)                 & $N/2$ (auto)             \\
    Inner epochs                & 1                            & 1                        \\
    Timestep fraction (PPO mask) & 0.99                        & 0.99                     \\
    KL $\beta$                  & 0.01                         & 0                        \\
    EMA on policy weights       & yes                          & yes                      \\
    Hardware                    & 1 node $\times$ 8 H200       & 2 nodes $\times$ 8 H200  \\
    Reported steps              & 1000--1500                   & 2000--2700               \\
    Save frequency              & every 50 steps               & every 50 steps           \\
    Eval frequency              & every 50 steps               & every 50 steps           \\
    \bottomrule
  \end{tabular}
\end{table}

\paragraph{DreamSync baseline.}
DreamSync iterations are pure supervised LoRA fine-tunes on filtered self-generated
data: for each iteration we generate $8$ candidates per prompt, score them with the
faithfulness reward and PickScore, keep the best per prompt, and fine-tune for $1000$
supervised steps with effective batch $256$ (SD3: $8\times 8 \times 4$;
FLUX: $4\times 8 \times 8$), AdamW, learning rate $3{\times}10^{-4}$, and EMA. Iteration
$N$ initializes from the LoRA of iteration $N-1$. We report $3$ iterations on each
backbone.

\paragraph{Reward weight conventions.}
All single-reward runs use weight $1.0$ on the lone reward. All combined runs use weight
$1.0$ on each of \emph{PickScore} and \emph{faithfulness} unless otherwise stated. Under
GDPO each weight multiplies the post-normalization advantage; under GRPO it multiplies
the raw reward.

\subsection{Reward models}
\label{app:training_details:rms}

\emph{PickScore}~\citep{pickscore} is a Bradley--Terry (BT) preference reward built on the
CLIP-H/14 backbone (OpenCLIP)~\citep{cherti2023reproducible} and fine-tuned on the Pick-a-Pic corpus of
${\sim}500$K crowd pairwise annotations of T2I generations collected from a
public web playground. Because annotators were asked for an open-ended 
verdict and the dual-tower CLIP backbone has
limited compositional reasoning, the resulting scalar in practice tracks
aesthetic appeal far more than fine-grained prompt fidelity.
\emph{HPSv3} (Human Preference Score v3)~\citep{hpsv3} is a more recent BT preference
model fine-tuned on top of a Qwen2.5-VL-7B backbone~\citep{qwen2_5_vl} using
the HPDv3 dataset, a curated corpus of ${\sim}1$M human pairwise
preferences sourced from a wider distribution of base models with
finer-grained annotation guidelines that explicitly cover both aesthetic
quality and prompt-faithfulness sub-criteria. The larger VLM backbone and
broader annotation rubric make HPSv3 a stronger holistic preference signal.
\emph{ImageReward}~\citep{imagereward} is a BT preference model based on a BLIP backbone and
trained on ${\sim}137$K expert-annotated rankings over generations sampled
from DiffusionDB; annotators score each image along prompt alignment,
fidelity, and harmlessness, so the resulting scalar emphasizes
text--image alignment more than the CLIP-tower PickScore.
\emph{UnifiedReward-2.0}~\citep{unifiedreward} is a VLM-based judge; we use the Qwen3.5-27B
variant of the v2.0 release. It is distilled into a single scalar from a
mix of pairwise and pointwise human-preference data spanning T2I,
image-to-image, and image-to-video tasks. It is designed as a
``one-reward-for-all'' proxy and is the only baseline that explicitly
tries to cover prompt-following and aesthetics jointly.
\subsection{Question-style ablation details}
\label{app:ablations}
\label{app:impl:generic-rules}

This subsection details the four question-style ablations introduced in
Section~\ref{sec:ablations:questions}. All four variants share the training
hyperparameters of Section~\ref{sec:experiments}, the GDPO summary
training objective with flat weight $1$ on each reward component, and the
same Qwen3.5-27B oneshot judge of Appendix~\ref{app:impl:vlm}. Only the
construction of the faithfulness signal differs across variants.

\paragraph{Ignore-dep.\ details.}
This variant uses the \emph{same} per-prompt question DAG produced by the
default Gemini-3-Pro decomposer (Appendix~\ref{app:impl:decompose}); only
the scoring rule changes. At scoring time the BFS-with-gating routine of
Section~\ref{sec:reward:design} is replaced by a flat scan: every question
is queried independently, parents do not zero out descendants, and the
faithfulness reward is the unweighted yes-ratio
$\frac{1}{|Q_f(p)|}\sum_q y_q$ over the entire faithfulness subgraph.
The VLM call count therefore goes \emph{up} by exactly the number of
questions that the default would have skipped via gating; this is a
pure scoring-rule ablation, not a question-set ablation.

\paragraph{Generic details.}
The \emph{Generic} variant replaces the prompt-specific decomposition with
a fixed list of $15$ prompt-agnostic faithfulness rules that every prompt
shares. The judge is queried with the same oneshot system prompt of
Appendix~\ref{app:impl:vlm}; only the question list changes. The rules are
reproduced verbatim below (CSV column \texttt{rules}, source
\path{flow_grpo/rules_general_faithfulness.csv}).

\begin{quote}\small
\begin{enumerate}\itemsep0em
  \item Does the image show the main subject or scene described in the prompt?
  \item Is the image overall relevant to the prompt?
  \item Are the key objects or entities mentioned in the prompt present?
  \item Are no important requested elements missing?
  \item Do the visible attributes of the main subjects match the prompt?
  \item Are important prompt-specific details correctly shown?
  \item Does the number of key objects or subjects match the prompt?
  \item Are the subjects performing the actions described in the prompt?
  \item Are the relationships between subjects consistent with the prompt?
  \item Is the spatial arrangement consistent with the prompt?
  \item Does the background or environment match the prompt?
  \item Is the location or setting consistent with the prompt?
  \item Does the time, weather, or season match the prompt, if specified?
  \item Does the visual style match the prompt, if specified?
  \item Is the image faithful to the prompt overall?
\end{enumerate}
\end{quote}

These rules are framed at a uniformly generic level---they reference ``the
prompt'', ``the main subjects'', ``key objects'' rather than specific
entities or relations. Because they cannot encode any prompt-specific
structure, the checklist reward in this variant collapses to a coarse
``is the image roughly faithful'' scalar.

\paragraph{Faith\,$+$\,Aesth details.}
The \emph{Faith\,$+$\,Aesth} variant uses both decomposition passes of
Appendix~\ref{app:impl:decompose}. For each prompt the offline pipeline
emits the faithfulness DAG \emph{and} the parallel aesthetics DAG
(rendering quality, color harmony, lighting, composition, overall
coherence); their union $Q_f(p) \cup Q_a(p)$ is the complete question set.
The dependency walk and parent-gating rule are unchanged from the default,
but the per-image reward is the yes-ratio over the \emph{full} union, not
just the faithfulness subgraph. PickScore is dropped from the reward
mixture, so the checklist is the only signal driving the policy. This
isolates whether checklist-style supervision can absorb the role normally
played by a BT aesthetic reward.

\paragraph{RubricRL details.}
The \emph{RubricRL} variant uses a different offline decomposer prompt
that emits a flat, dependency-free rubric mixing faithfulness and
rendering-quality items. Every item is a root question
(\texttt{depends\_on}: $[]$), and the per-image reward is the unweighted
yes-ratio over the rubric, queried once per image in the same oneshot
mode. The decomposer system prompt is reproduced verbatim below.

\begin{Verbatim}
You are a rubric generation model for text-to-image evaluation.

Task:
Given one text-to-image prompt, extract the evaluation questions that a
careful human judge would use to determine whether a generated image
truly satisfies the prompt AND is aesthetically pleasing.

Goal:
Produce a question list that is:
- prompt-adaptive
- decomposable
- atomic
- visually checkable
- suitable for binary scoring (yes/no, pass/fail)

Instructions:
1. Read the prompt carefully and identify visually verifiable
   requirements.
2. Convert them into short, independent evaluation questions.
3. Cover the most important dimensions when relevant:
   - object count
   - object identity
   - attribute accuracy (color, material, texture, size)
   - action / pose
   - spatial relations / placement
   - OCR / visible text fidelity
   - scene coherence / composition
   - style consistency
   - aesthetic / image quality (rendering quality, lighting, color
     harmony)
   - special constraints such as monochrome, color palette, lighting,
     era, material, etc.
4. Do not include duplicate or overlapping questions.

Output format:
Return ONLY a valid JSON array of objects. Each object has:
  - "id": integer starting from 0
  - "question": one atomic yes/no question about the image

No explanation, no markdown fences. ONLY the JSON array.

Example for "A red cat sitting on a blue chair":
[
  {"id": 0, "question": "Is there a cat in the image?"},
  {"id": 1, "question": "Is the cat red?"},
  {"id": 2, "question": "Is there a chair in the image?"},
  {"id": 3, "question": "Is the chair blue?"},
  {"id": 4, "question": "Is the cat sitting on the chair?"},
  {"id": 5, "question": "Is the cat rendered with proper anatomy and
   realistic fur texture?"},
  {"id": 6, "question": "Does the chair have clean lines and convincing
   material texture?"},
  {"id": 7, "question": "Is the overall image composition well-balanced?"},
  {"id": 8, "question": "Is the lighting across the scene consistent and
   visually appealing?"},
  {"id": 9, "question": "Do the colors in the image look harmonious and
   natural?"}
]
\end{Verbatim}

Note that the output schema requested above only carries \texttt{id}
and \texttt{question} fields---no \texttt{depends\_on} field is
emitted by the decomposer---so the resulting rubric is flat by
construction; in our post-processing we materialise this as a
\texttt{depends\_on: []} entry on every item, and we verified
empirically across the full $10{,}000$-prompt rubric training set
that $0/155{,}434$ items carry a non-empty parent list. Items per
prompt are also higher than the example suggests in practice: mean
$15.5$, median $15$, range $[0,68]$ across the $10{,}000$-prompt
training set. The rubric mixes faithfulness questions (object
existence, attributes, spatial relations) with rendering-quality
questions (anatomy, composition, lighting, color harmony) in a single
flat list. Compared to our default, RubricRL therefore differs in two
coupled ways: it removes the dependency structure (similar to
\emph{Ignore-dep.}) \emph{and} pushes aesthetic items into the same
checklist (similar to \emph{Faith\,$+$\,Aesth}). It approximates the
rubric style used in prior rubric-based RLHF work for T2I and serves
as a unified-checklist baseline against our structured faithfulness
signal plus a separate BT aesthetic reward.

\subsection{DreamSync Iterative SFT Baseline}
\label{app:dreamsync}

DreamSync is the iterative-SFT baseline referenced in Section~\ref{sec:experiments}. It alternates an outer \emph{generate-and-filter} step, which constructs a reward-selected supervised set under the current policy, with a \emph{LoRA fine-tune} step on that set. Iteration $0$ starts from the base model with a fresh LoRA. For iteration $N{>}0$, we initialize from iteration $N{-}1$'s LoRA checkpoint, so the policy improves cumulatively. We run three iterations on each backbone.

\paragraph{Generate-and-filter.}
At the start of each iteration, the current policy generates $K{=}8$ candidate images for every prompt in the same 10k training set from Section~\ref{sec:arena-t2i}, yielding $80{,}000$ candidates per iteration. Sampling uses the same flow-matching scheduler and backbone-specific settings as the RL runs: $T{=}40$ denoising steps and CFG~$=4.5$ at $1024{\times}1024$ for SD3.5-M, and $T{=}50$ denoising steps and CFG~$=3.5$ at $1024{\times}1024$ for FLUX.1-dev.

Each candidate is scored independently by PickScore, used as the aesthetic signal, and by our Qwen3.5-27B checklist reward in oneshot mode, used as the faithfulness signal. We then apply a median prefilter: a candidate is retained only if its PickScore is at or above the iteration-level median PickScore and its faithfulness yes-ratio is at or above the iteration-level median yes-ratio. Among the surviving candidates for each prompt, we select a single supervised target using the lexicographic key
\[
(\text{faithfulness}, \text{PickScore}),
\]
breaking ties by PickScore. Prompts for which all candidates are filtered out are omitted from that iteration's supervised set.

\paragraph{LoRA fine-tune.}
The retained $(\text{prompt}, \text{image})$ pairs are used as ground truth for a flow-matching SFT objective:
\[
\mathcal{L}_{\text{SFT}}
=
\mathbb{E}\left[
\left\|
v_\theta(x_t,t,p) - (\epsilon - x_0)
\right\|_2^2
\right],
\]
where $x_0$ is the latent of the retained image, $\epsilon$ is the noise sample, and $\epsilon - x_0$ is the standard flow-matching velocity target.

Optimization mirrors the RL setting. We use AdamW with learning rate $3{\times}10^{-4}$, $(\beta_1,\beta_2)=(0.9,0.999)$, weight decay $10^{-4}$, and train for $1000$ supervised steps per iteration. We maintain an EMA of the LoRA weights with decay $0.9$, updated every $8$ steps. SD3.5-M uses LoRA rank $r{=}32$, $\alpha{=}64$, applied to the $8$ attention Q/K/V/output projections, with effective batch size $256$ ($8$ GPUs $\times$ batch $8$ $\times$ $4$ gradient accumulation) in \texttt{fp16}. FLUX.1-dev uses rank $r{=}64$, $\alpha{=}128$, applied to $12$ attention and feedforward modules, with effective batch size $256$ ($8$ GPUs $\times$ batch $4$ $\times$ $8$ gradient accumulation) in \texttt{bf16}. The training resolution is $512{\times}512$, matching the RL runs.

\paragraph{Iteration loop.}
Iteration $0$ starts from the base model with a fresh LoRA. For iteration $N{>}0$, we load the LoRA from iteration $N{-}1$'s \texttt{checkpoint-1000} as the initialization of a new LoRA with the same shape. LoRAs are not merged into the base weights, so all iterations train the same number of parameters. Each iteration re-runs generate-and-filter with the updated policy, allowing the supervised set to track the model's own evolving generation distribution. This iterative refresh is the key distinction from a single-pass best-of-$K$ SFT baseline.

\section{Additional Results}
\label{app:results}

This appendix collects results that did not fit in the main paper.

\subsection{Training curves}
\label{app:results:curves}

Figure~\ref{fig:training-curves} (single-reward fine-tunes on SD3.5-M) and
Figure~\ref{fig:training-curves-combined} (FLUX.1-dev with the combined
Faith\,$+$\,Pick run) plot the relative change of three held-out eval rewards
against training step. Two qualitative patterns are visible: BT-only
fine-tunes plateau early on the checklist reward (sometimes regressing below
the base level) while continuing to climb on PickScore; combined GDPO runs
continue to gain on the checklist reward without dropping PickScore.

\subsection{Cross-seed stability of matrix~1 win rates}
\label{app:results:rowmean}

\begin{figure}[t]
  \centering
  \begin{subfigure}{0.49\linewidth}
    \centering
    \includegraphics[width=\linewidth]{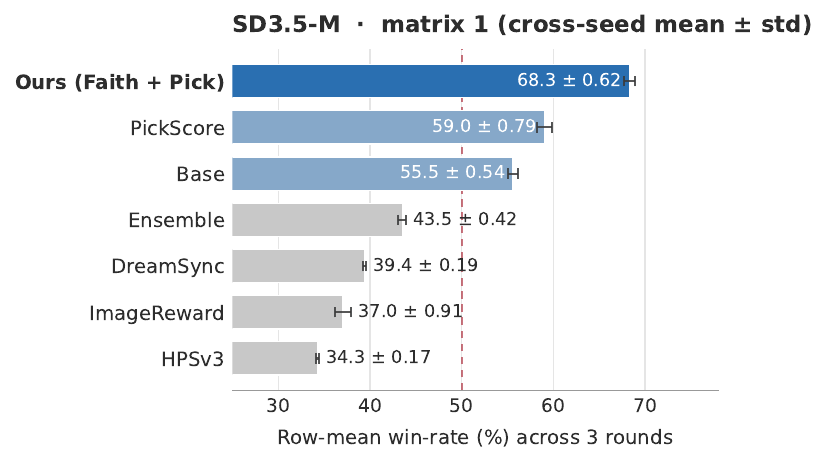}
    \caption{SD3.5-M row-mean $\pm$ std (3 rounds)}
    \label{fig:matrix-sd3-rowmean}
  \end{subfigure}
  \hfill
  \begin{subfigure}{0.49\linewidth}
    \centering
    \includegraphics[width=\linewidth]{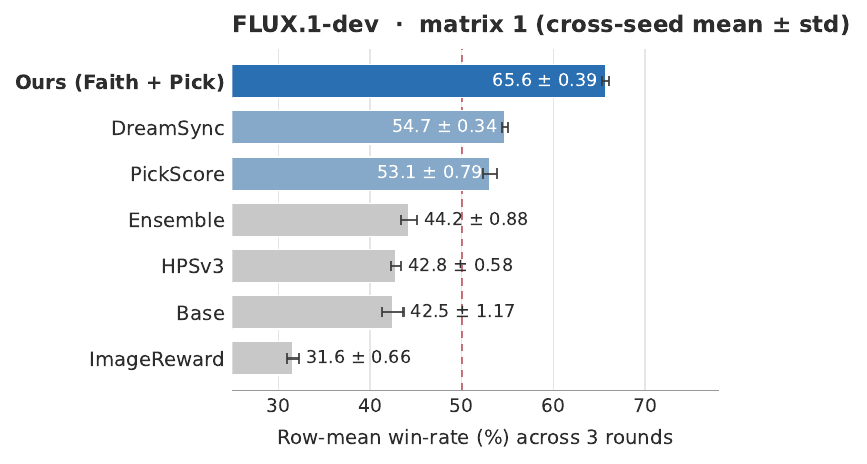}
    \caption{FLUX.1-dev row-mean $\pm$ std (3 rounds)}
    \label{fig:matrix-flux-rowmean}
  \end{subfigure}
  \caption{Cross-seed row-mean win rates for matrix~1
  (Figure~\ref{fig:matrix-baseline}). Each model's bar is its row-mean
  win rate against the rest of the matrix, averaged over $3$
  random-seed rounds (\texttt{image\_idx}~$0/1/2$); error bars show
  the across-round standard deviation. The $50\%$ reference line
  marks the no-effect threshold. Faith\,$+$\,Pick (highlighted) is
  the top row on both backbones, and the ranking is stable across
  rounds.}
  \label{fig:matrix-baseline-rowmean}
\end{figure}

To quantify sampling noise on the matrix~1 head-to-head numbers in
Section~\ref{sec:results:main}, we regenerate the entire matrix with
two extra random seeds (\texttt{image\_idx}~$1, 2$ in addition to
\texttt{image\_idx}${=}0$) and compute each model's row-mean win rate
against the rest of the matrix in every round.
Figure~\ref{fig:matrix-baseline-rowmean} reports the mean and standard
deviation of these row-mean win rates across the three rounds. The
maximum cross-round standard deviation is $0.91$~pp on SD3.5-M
(ImageReward) and $1.17$~pp on FLUX.1-dev (Base), an order of
magnitude smaller than the gaps between models. The matrix~1 ranking
is therefore stable across seeds, and the headline result that
Faith\,$+$\,Pick is the top row on both backbones is not driven by a
favourable single-seed draw.

\subsection{GDPO sub-mode ablation}
\label{app:results:gdpo}
\label{sec:method:submodes}

The dependency-aware faithfulness reward of Section~\ref{sec:reward}
exposes both a summary score $s_f(i,j) \in [0, 1]$ and a per-question
vector $\mathbf{y}(i,j) \in \{0, 1\}^{n_i}$, where $n_i = |Q_f(p_i)|$
is the number of faithfulness questions for prompt $i$. Given the
GRPO/GDPO formulation in Section~\ref{sec:method}, three ways of
plugging this signal into the policy gradient are natural; we ablate
all three on SD3.5-M.

\emph{Vanilla.} The naive GRPO baseline: take the simple weighted sum
$r(i,j) = w_{\text{BT}} \cdot r_{\text{BT}}(i,j) + w_{\text{faith}}
\cdot s_f(i,j)$ and apply Eq.~\eqref{eq:grpo} with no per-reward
normalization. This is the source of the signal collapse described in
Section~\ref{sec:method}.

\emph{Summary.} A GDPO sub-mode that treats the faithfulness reward as
a single signal $r_{\text{faith}}(i,j) = s_f(i,j)$ and applies
Eq.~\eqref{eq:gdpo} unchanged. Faithfulness contributes one normalized
advantage per rollout, on equal footing with each other reward.

\emph{GRPO-vanilla.} A GDPO sub-mode that treats each question as its
own reward. For prompt $i$ and rollout $j$, define
$A_{\text{faith}}^{\text{grpo-vanilla}}(i,j) = \sum_{q=1}^{n_i}
(y_q(i,j) - \mu_{q,i}) / (\sigma_{q,i} + \varepsilon)$. Each per-question
score is normalized within the rollout group, then summed. Because the
sum has $n_i$ terms, this mode scales the effective faithfulness weight
with the number of questions per prompt. The weight $w_{\text{faith}}$
in Eq.~\eqref{eq:gdpo} is multiplied on top.

The combined-reward main result (Faith\,$+$\,Pick) reported in
Section~\ref{sec:results:main} is trained under \emph{Summary}. To
isolate the contribution of the combination strategy, we run all three
sub-modes under identical conditions on SD3.5-M---same backbone, same
PickScore $+$ checklist mixture at flat weight $1$, same 1000 training
steps---and compare them pairwise on a separate $1{,}000$-prompt
held-out evaluation set drawn from the same arena source pool.

\begin{figure}[t]
  \centering
  \includegraphics[width=0.5\linewidth]{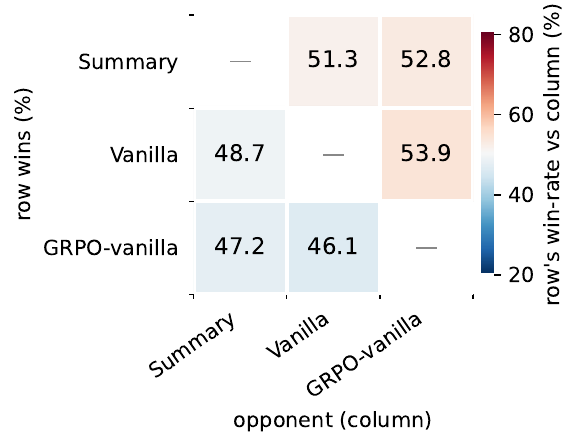}
  \caption{GDPO sub-mode ablation on a separate $1{,}000$-prompt
  held-out evaluation set (SD3.5-M, ckpt-1000). All three runs use
  PickScore $+$ our checklist reward at flat weight $1$ each; only the
  combination strategy varies.}
  \label{fig:matrix-gdpo-medium}
\end{figure}

On the holistic MMRB2 verdict, the three strategies order cleanly:
$\textit{Summary} > \textit{Vanilla} > \textit{GRPO-vanilla}$.
\emph{Summary} beats the \emph{Vanilla} weighted-sum baseline $51.3\%$ to
$48.7\%$ and beats \emph{GRPO-vanilla} $52.8\%$ to $47.2\%$, while
\emph{Vanilla} in turn beats \emph{GRPO-vanilla} $53.9\%$ to $46.1\%$.
The per-axis breakdown in
Appendix~\ref{app:results:per-axis} is more nuanced: on the
aesthetics axis the holistic ordering holds (Summary $51.4\%$ vs
Vanilla, $52.7\%$ vs GRPO-vanilla), while on the faithfulness axis
\emph{Vanilla} actually beats Summary $57.2\%$ to $42.8\%$ and beats
GRPO-vanilla $52.6\%$ to $47.4\%$. \emph{Summary's holistic win is
therefore driven by aesthetics, not faithfulness.}
We use \emph{Summary} as the default GDPO sub-mode for every combined
result elsewhere in the paper because the holistic verdict is what the
main MMRB2 evaluation tracks; the broadcast variant
(\emph{GRPO-vanilla}) under-performs both axes at this scale.

\subsection{Per-axis breakdown of MMRB2 win-rates}
\label{app:results:per-axis}

The MMRB2 judge writes free-text reasoning for each evaluation criterion
and emits a single overall \texttt{better\_response} verdict. The
matrices in Section~\ref{sec:results:main} report this overall verdict.
For finer-grained analysis we extract two axes from the same pairwise
JSONs: \emph{aesthetics}, taken directly from the integer \texttt{score}
(the criterion explicitly described in the rubric as ``general technical
and aesthetic quality, realism, coherence''); and \emph{faithfulness},
inferred per-pair from the per-criterion reasoning prose for
\texttt{faithfulness\_to\_prompt} via regex/keyword heuristics. The
heuristic parser leaves $\sim 30\!-\!40\%$ of judgements unparsed; the
per-cell denominator in each figure shows how many pairs voted
decisively, so the reader can spot small samples.

\paragraph{Baseline matrices: Faith\,$+$\,Pick wins more on the
faithfulness axis than on aesthetics for every BT baseline.}
On both backbones, our combined Faith\,$+$\,Pick run's
\emph{faithfulness} win-rate matches or exceeds its \emph{aesthetics}
win-rate against every BT preference baseline. On SD3.5-M:
$72.4\%$ vs.\ $59.7\%$ against PickScore, $86.2\%$ vs.\ $77.2\%$
against HPSv3, $81.1\%$ vs.\ $77.5\%$ against ImageReward,
$80.5\%$ vs.\ $71.9\%$ against the ensemble. On FLUX.1-dev: $72.6\%$
vs.\ $61.9\%$ against PickScore, $74.3\%$ vs.\ $70.2\%$ against
HPSv3, and $74.4\%$ vs.\ $68.6\%$ against the ensemble. The
DreamSync direction is the exception: DreamSync filters its supervised
set partly on the same checklist signal, so it is competitive on
faithfulness ($63.7\%$ on SD3, $57.4\%$ on FLUX) and Faith\,$+$\,Pick's
edge over it is mostly aesthetic ($74.2\%$ on SD3, $58.9\%$ on FLUX).

\begin{figure}[t]
  \centering
  \begin{subfigure}{0.49\linewidth}
    \centering
    \includegraphics[width=\linewidth]{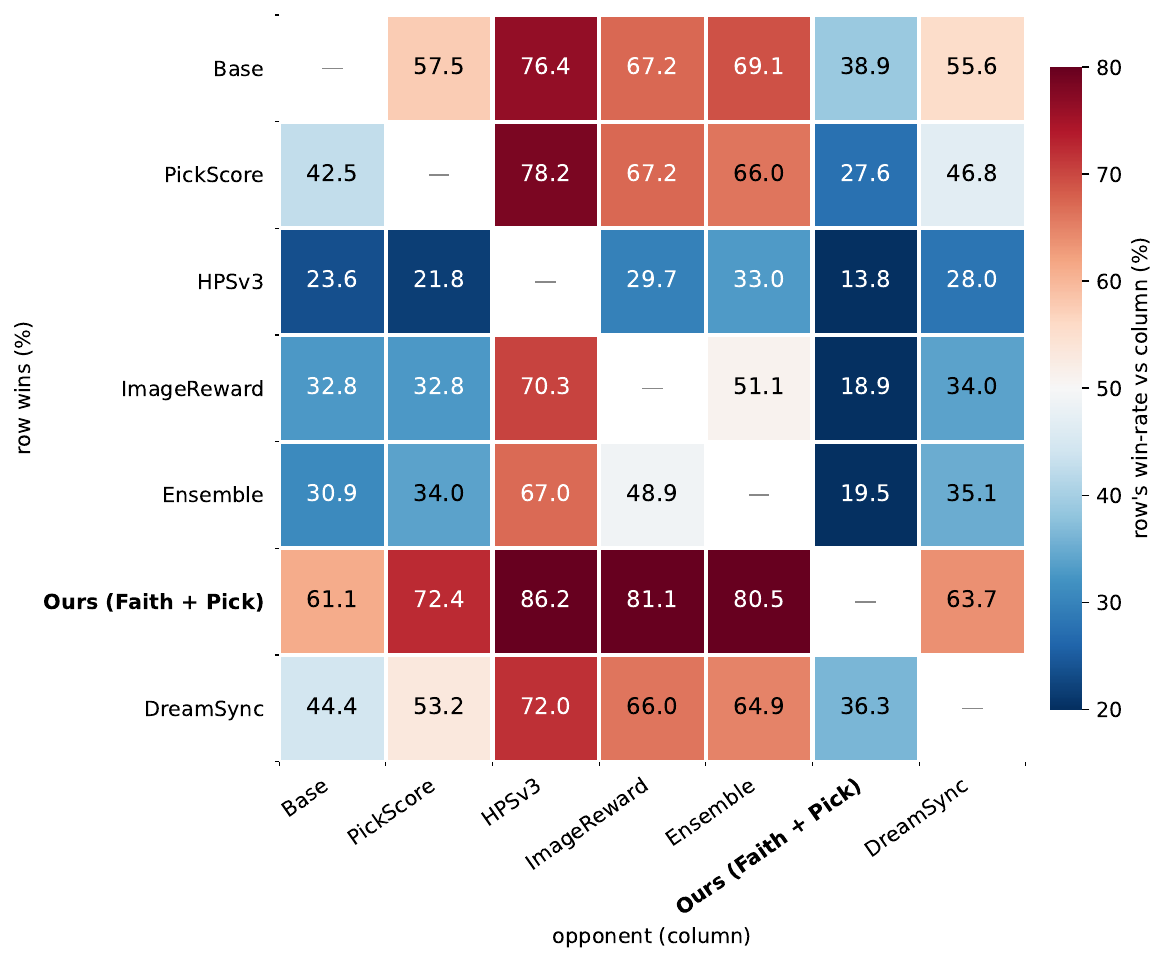}
    \caption{SD3.5-M, faithfulness axis}
  \end{subfigure}
  \hfill
  \begin{subfigure}{0.49\linewidth}
    \centering
    \includegraphics[width=\linewidth]{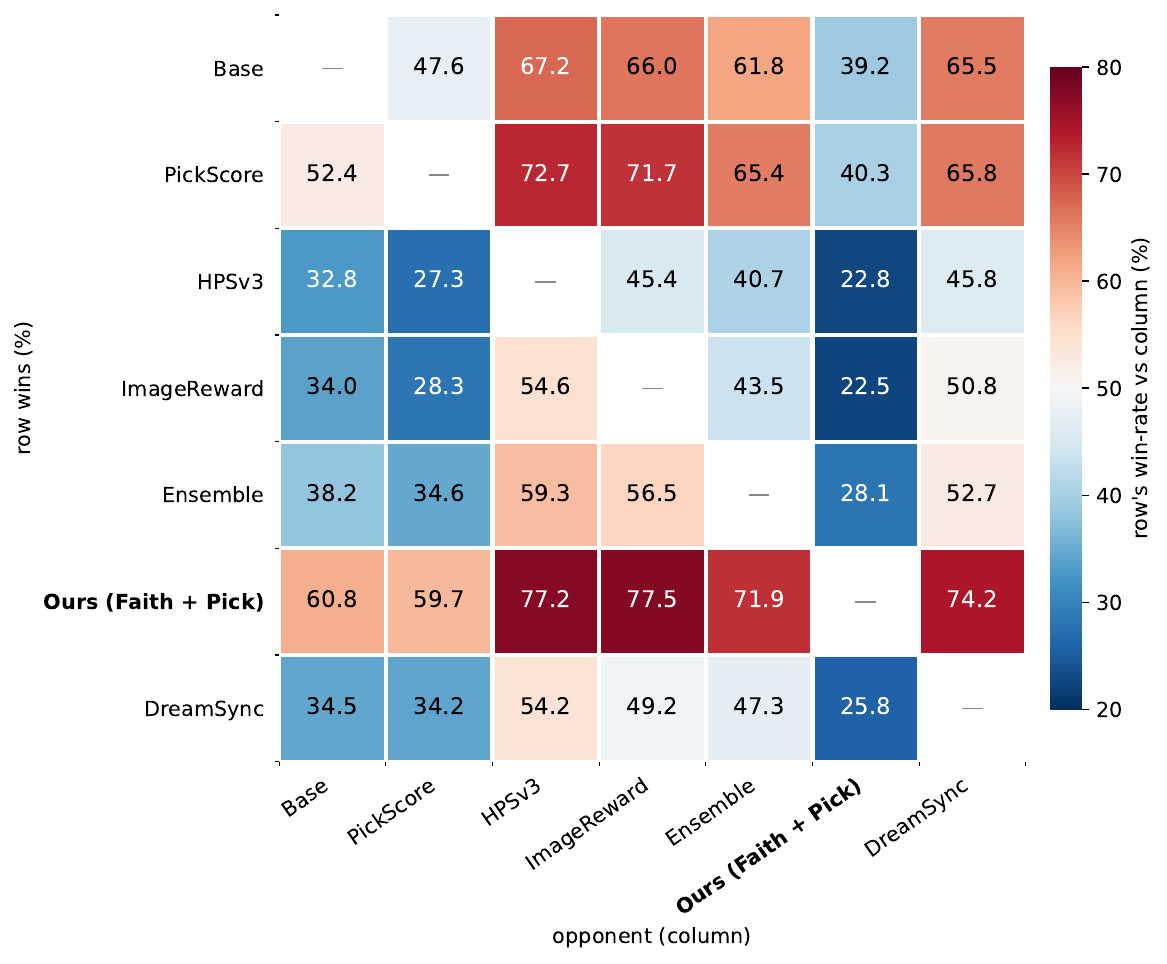}
    \caption{SD3.5-M, aesthetics axis}
  \end{subfigure}\\
  \begin{subfigure}{0.49\linewidth}
    \centering
    \includegraphics[width=\linewidth]{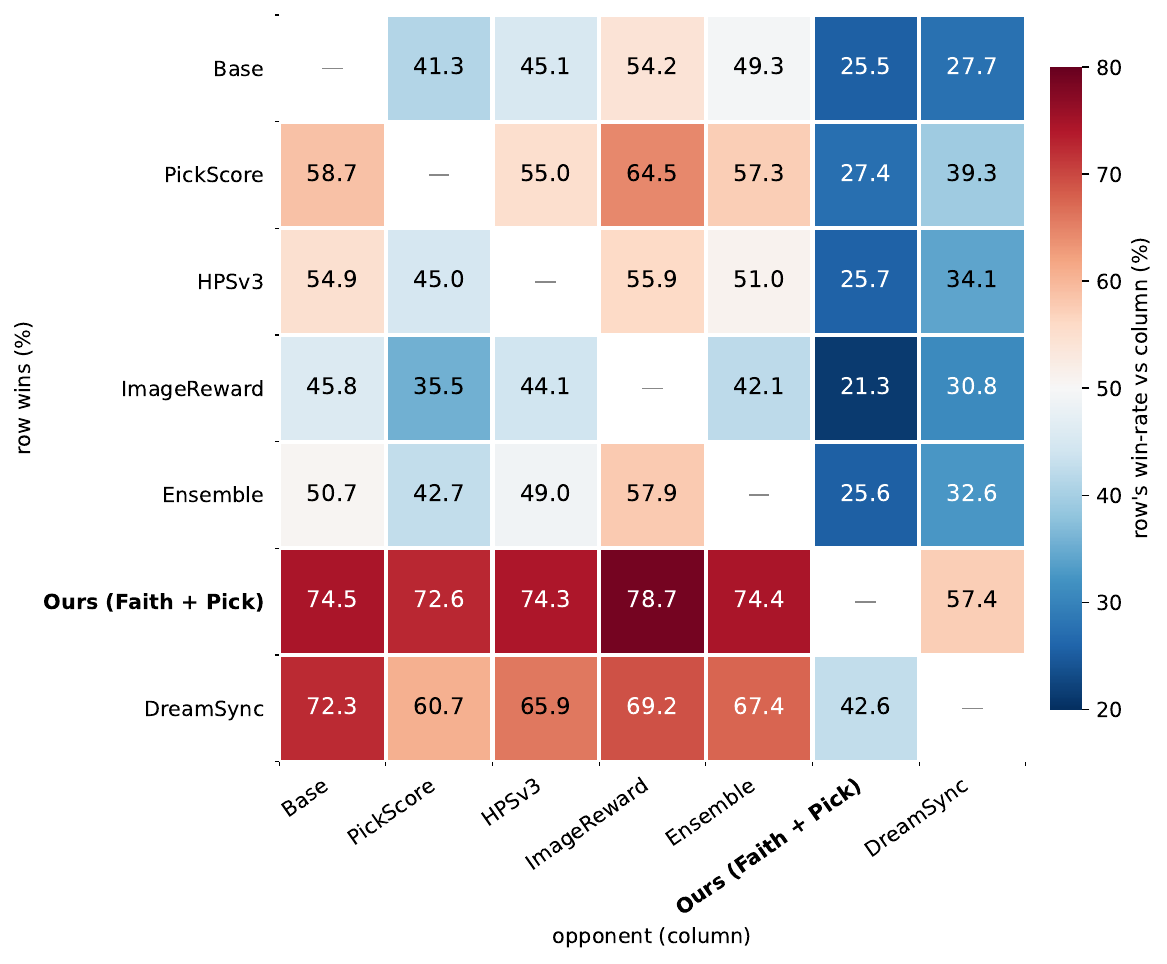}
    \caption{FLUX.1-dev, faithfulness axis}
  \end{subfigure}
  \hfill
  \begin{subfigure}{0.49\linewidth}
    \centering
    \includegraphics[width=\linewidth]{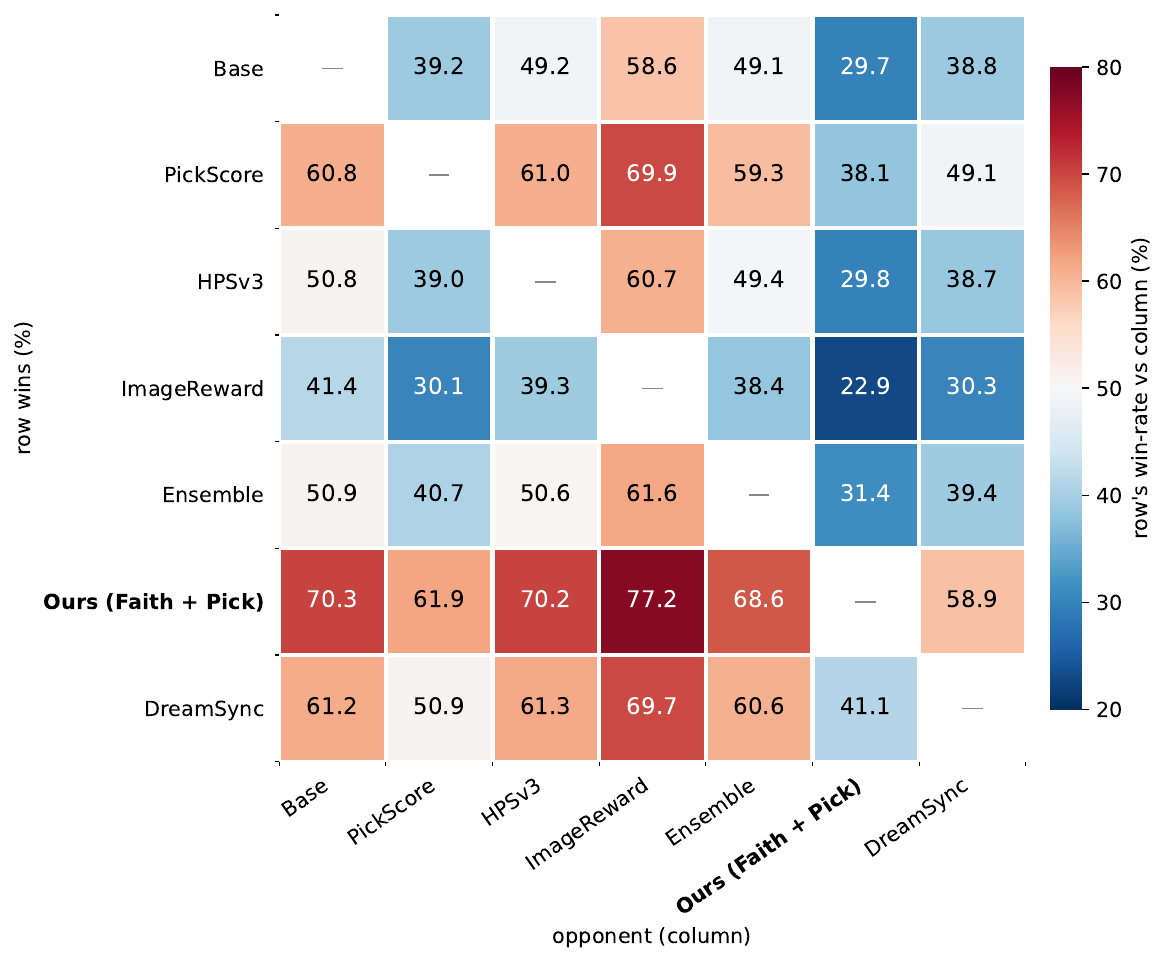}
    \caption{FLUX.1-dev, aesthetics axis}
  \end{subfigure}
  \caption{Per-axis baseline matrices on the 1k test set. Each cell is
  the row's win-rate against the column on a single criterion; the
  faithfulness axis is parsed heuristically from the judge's
  per-criterion reasoning, the aesthetics axis is the integer
  \texttt{score} ($4$--$6 \to$ A wins). The Faith\,$+$\,Pick row's
  faithfulness margins exceed its aesthetics margins for every BT
  preference baseline.}
  \label{fig:matrix-axis-baseline}
\end{figure}

\paragraph{GDPO sub-mode ablation: Summary's win is aesthetics-only.}
Decomposing Figure~\ref{fig:matrix-gdpo-medium} by axis flips the
ordering on faithfulness. \emph{Summary} keeps its narrow win on
aesthetics ($51.4\%$ vs Vanilla, $52.7\%$ vs GRPO-vanilla), but on
faithfulness it is the \emph{worst} of the three: \emph{Vanilla} beats
Summary $57.2\%$ to $42.8\%$, and \emph{GRPO-vanilla} also beats
Summary $57.2\%$ to $42.8\%$. \emph{Vanilla} edges \emph{GRPO-vanilla}
on both axes ($53.9\%$ aesthetics, $52.6\%$ faithfulness). The
holistic ranking $\textit{Summary} > \textit{Vanilla} >
\textit{GRPO-vanilla}$ is therefore driven entirely by the aesthetics
axis; on faithfulness, the ranking inverts to
$\textit{Vanilla} > \textit{GRPO-vanilla} > \textit{Summary}$.

\paragraph{Rules ablation: Faith\,$+$\,Pick trades faithfulness for aesthetics.}
The headline rules-ablation matrix (Figure~\ref{fig:matrix-rules-ablation})
shows Faith\,$+$\,Pick winning every cell of the top row. Decomposing
by axis paints a more nuanced picture:
on \emph{aesthetics} Faith\,$+$\,Pick wins everywhere ($54.2\%$ over
Ignore-dep., $53.4\%$ over Faith\,$+$\,Generic, $51.7\%$ over
Faith\,$+$\,Aesth, $53.0\%$ over RubricRL), but on \emph{faithfulness}
alone the ablations that invest more checklist budget in faithfulness
questions \emph{beat} Faith\,$+$\,Pick: Faith\,$+$\,Aesth wins
$60.3\%$ to $39.7\%$, RubricRL wins $54.8\%$ to $45.2\%$. Faith\,$+$\,Pick
still beats Faith\,$+$\,Generic on faithfulness ($62.1\%$, the
prompt-specific decomposition matters) and slightly beats Ignore-dep.\
($52.5\%$, the dependency walk matters). The honest reading is that
\emph{Faith\,$+$\,Pick is a sweet-spot trade-off, not a strict
faithfulness maximizer}: Faith\,$+$\,Aesth and RubricRL push
faithfulness higher at the cost of aesthetics, and lose the holistic
verdict; Faith\,$+$\,Pick gives up some faithfulness in exchange for a
BT aesthetic signal that pulls the holistic verdict back ahead.

\subsection{Human study: validating the matrix~1 judge}
\label{app:results:human-study}

The matrix~1 head-to-head numbers in Section~\ref{sec:results:main}
are produced by a Gemini-3-flash judge under the MMRB2 rubric. To
verify that this judge tracks human preference rather than its own
biases, we ran a parallel human-vote study using the same matrix~1
SD3.5-M pairings.

\paragraph{Protocol.}
We anchor on our combined \emph{Faith\,$+$\,Pick} run (the
\texttt{rubricfaith\_sd3} checkpoint at step~$1000$) and pair it
against each of the six other models in matrix~1: \emph{Base},
\emph{PickScore}, \emph{HPSv3}, \emph{ImageReward}, \emph{DreamSync},
and the \emph{$4$-reward ensemble}. For every (anchor, opponent)
pair we draw $300$ prompts from the $1{,}000$-prompt MMRB2 evaluation
set, stratified $100/100/100$ across the
\emph{easy}/\emph{medium}/\emph{hard} difficulty bands, yielding
$1{,}800$ voting rows split into three self-contained HTML pages.
Pre-rendered $512{\times}512$ images are embedded as base-64 JPEGs;
each row shows the two images side by side with random A/B
assignment, and the voter is forced to pick A or B (no ties; the
VLM-judge verdict is never shown). Voter onboarding goes through a
modal that hashes the voter's name (FNV-1a $\to$ mulberry32) and
shuffles the row order with a per-voter Fisher--Yates pass, so
different voters see a different first-$50$ batch and the workload
spreads naturally across the full set without coordination.
Per-voter JSON files record
\texttt{(prompt\_id, difficulty, anchor, opponent, anchor\_side,
vote, winner)}. After the headline run, the \emph{Base-SD3} cell was
the closest peer to chance, so we ran a focused follow-up
mini-study---a single-pair HTML with a fresh $300$-prompt
stratified-random draw at seed~$142$ (vs.\ $42$ in the headline
run)---to thicken that cell with $\sim$$3{\times}$ the per-pair
sample size of the original.

\paragraph{Results.}
Table~\ref{tab:human-study} reports the aggregated human votes for
all six opponent cells, including the follow-up Base-SD3 votes
($n{=}760$ total for that cell after combining the original
$264$ votes with the $496$ follow-up votes). Faith\,$+$\,Pick wins
\textbf{$1218$ of $1899$ pairs ($64.1\%$ overall)} against the six
opponents combined, and beats every individual opponent. The
ImageReward cell is the most decisive ($95.5\%$ for Faith\,$+$\,Pick),
consistent with ImageReward being the weakest BT preference
fine-tune in matrix~1; the Base-SD3 cell is the closest ($55.9\%$
after the follow-up), again consistent with the VLM-judge
assessment that the SD3 base model is itself a strong starting
point on the easy 1k test set.

\begin{table}[t]
  \centering
  \footnotesize
  \setlength{\tabcolsep}{4pt}
  \begin{tabular}{lrrrrr}
  \toprule
  \textbf{Faith\,$+$\,Pick vs.}
              & \textbf{Anchor wins} & \textbf{Opponent wins} & \textbf{$N$}
              & \textbf{Human win-rate}
              & \textbf{Gemini-3-flash win-rate} \\
  \midrule
  \emph{ImageReward}            & $213$  & $\phantom{0}10$  & $223$  & $\mathbf{95.5\%}$ & $77.5\%$ \\
  \emph{$4$-reward ensemble}    & $170$  & $\phantom{0}72$  & $242$  & $70.2\%$          & $71.7\%$ \\
  \emph{DreamSync}              & $137$  & $\phantom{0}74$  & $211$  & $64.9\%$          & $74.2\%$ \\
  \emph{HPSv3}                  & $119$  & $\phantom{0}70$  & $189$  & $63.0\%$          & $77.3\%$ \\
  \emph{PickScore}              & $154$  & $120$            & $274$  & $56.2\%$          & $59.8\%$ \\
  \emph{Base-SD3}$^\dagger$     & $425$  & $335$            & $760$  & $55.9\%$          & $60.9\%$ \\
  \midrule
  \textbf{Overall}              & $\mathbf{1218}$ & $\mathbf{681}$ & $\mathbf{1899}$ & $\mathbf{64.1\%}$ & --- \\
  \bottomrule
  \end{tabular}
  \caption{\textbf{Human-vote validation of the matrix~1 SD3.5-M
  Gemini-3-flash judge.} Anchor is Faith\,$+$\,Pick
  (\texttt{rubricfaith\_sd3} ckpt-$1000$); the six rows enumerate the
  other matrix~1 models. Human votes are aggregated across all
  voters who completed at least the first-$50$ shuffled batch.
  ${}^\dagger$ The Base-SD3 cell combines the original $264$ headline
  votes with $496$ additional votes from a focused follow-up
  mini-study at seed~$142$ (see protocol). The Gemini-3-flash column
  is the corresponding cell of Figure~\ref{fig:matrix-baseline}.
  Across all six cells the human and VLM-judge orderings agree on
  the sign of the Faith\,$+$\,Pick advantage; Faith\,$+$\,Pick wins
  every cell under both judges.}
  \label{tab:human-study}
\end{table}

\paragraph{Comparison to the VLM judge.}
On every one of the six cells, both the human voters and the
Gemini-3-flash judge place Faith\,$+$\,Pick above the opponent
(every cell is $>50\%$ in both columns). The two columns rank-correlate
at \textbf{Spearman $\rho = 0.66$} across the six cells; the absolute
gap between them ranges from $0.4$~pp (PickScore) to $18.0$~pp
(ImageReward), with humans pushing the ImageReward cell higher
(possibly because ImageReward generations have visible CLIP-style
artefacts that humans down-weight more aggressively than the VLM
judge). The bottom-line conclusion is the same under either judge:
on the six matrix~1 SD3.5-M opponents, Faith\,$+$\,Pick is the top
row.



\subsection{Limitations}
\label{app:limitations}
\label{sec:conclusion:limitations}

The faithfulness reward depends on frozen VLM/VLM components (the
offline Gemini-3-Pro decomposer, the Qwen3.5-27B reward judge inside
the training loop, and the gemini-3-flash judge used for the
closed-source leaderboard); the decomposer-robustness check in
Appendix~\ref{app:bench:hard}
(Table~\ref{tab:decomposer-robustness}) shows the leaderboard ranking
is stable under a second decomposer (GPT-5.4), but we do not measure
sensitivity to swapping the in-loop judge during RL training itself.
We also do not test the reward on other image-RL methods beyond
Flow-GRPO/GDPO.


%

\end{document}